\documentclass[10pt,twocolumn,letterpaper]{article}

\usepackage{iccv}
\usepackage{times}
\usepackage{epsfig}
\usepackage{graphicx}
\usepackage{amsmath}
\usepackage{amssymb}

\usepackage{multirow}
\usepackage{float}
\usepackage{subfig}
\usepackage{tabularx} 

\usepackage{xcolor}

\usepackage{pifont}

\newcommand{\xmark}{\ding{55}}

\usepackage[breaklinks=true,bookmarks=false]{hyperref}

\iccvfinalcopy 

\def\iccvPaperID{****} 

\ificcvfinal\fi

\begin{document}
	
	\title{Video Contrastive Learning with Global Context}

	\author{
		Haofei Kuang\thanks{Work done during an internship at Amazon.}~\textsuperscript{\rm 1,3}, Yi Zhu\textsuperscript{\rm 2}, Zhi Zhang\textsuperscript{\rm 2}, Xinyu Li\textsuperscript{\rm 2}, Joseph Tighe\textsuperscript{\rm 2}, \\Sören Schwertfeger\textsuperscript{\rm 3}, Cyrill Stachniss\textsuperscript{\rm 1}, Mu Li\textsuperscript{\rm 2} \\
		\textsuperscript{\rm 1} University of Bonn \,
		\textsuperscript{\rm 2} Amazon Web Services \,
		\textsuperscript{\rm 3} ShanghaiTech University \\
	}
	
	
	\maketitle
	\ificcvfinal\thispagestyle{empty}\fi
	
	\begin{abstract}
		Contrastive learning has revolutionized self-supervised image representation learning field, and recently been adapted to video domain. One of the greatest advantages of contrastive learning is that it allows us to flexibly define powerful loss objectives as long as we can find a reasonable way to formulate positive and negative samples to contrast. However, existing approaches rely heavily on the short-range spatiotemporal salience to form clip-level contrastive signals, thus limit themselves from using global context.
		In this paper, we propose a new video-level contrastive learning method based on segments to formulate positive pairs. Our formulation is able to capture global context in a video, thus robust to temporal content change. We also incorporate a temporal order regularization term to enforce the inherent sequential structure of videos. 
		Extensive experiments show that our video-level contrastive learning framework (VCLR) is able to outperform previous state-of-the-arts on five video datasets for downstream action classification, action localization and video retrieval. Code is available at \url{https://github.com/amazon-research/video-contrastive-learning}.
	\end{abstract}
	
	\section{Introduction}
	\label{sec:introduction}
	Self-supervised video representation learning has garnered great attention in the last several years. The allure of this paradigm is the promise of annotation-free ground-truth to ultimately learn a superior data representation. Initial attempts to improve the quality of learned representations were via crafting various pretext tasks \cite{vondrick_cvpr2016_anticipating,diba_iccv2019_dynamoNet,misra_eccv2016_shuffle,xu_cvpr2019_vcop,jing_arxiv2018_videoRotation,kim_aaai2019_cubicPuzzle,wang_cvpr2019_statistics,benaim_cvpr2020_speednet,wang_cvpr2019_cycleTime}. Recently, contrastive learning methods \cite{chen_arxiv2020_mocov2,chen_arxiv2020_simclrv2,grill_arxiv2020_byol,caron_nips2020_swav} based on instance discrimination \cite{wu_cvpr2018_instdisc} have revolutionized the self-supervised image representation learning field. They have consistently outperformed their supervised counterparts on downstream tasks like image classification, object detection and segmentation. Hence, several recent works have started to consider contrastive learning in the video domain \cite{sun_arxiv2019_cbt,han_iccvw2019_dpc,han_eccv2020_memdpc,wang_eccv2020_pace,tian_eccv2020_cmc,yang_arxiv2020_vthcl,yao_aaai2021_seco,qian_cvpr2021_cvrl}.
	
	\begin{figure}[t]
		\begin{center}
			\includegraphics[width=1  \columnwidth]{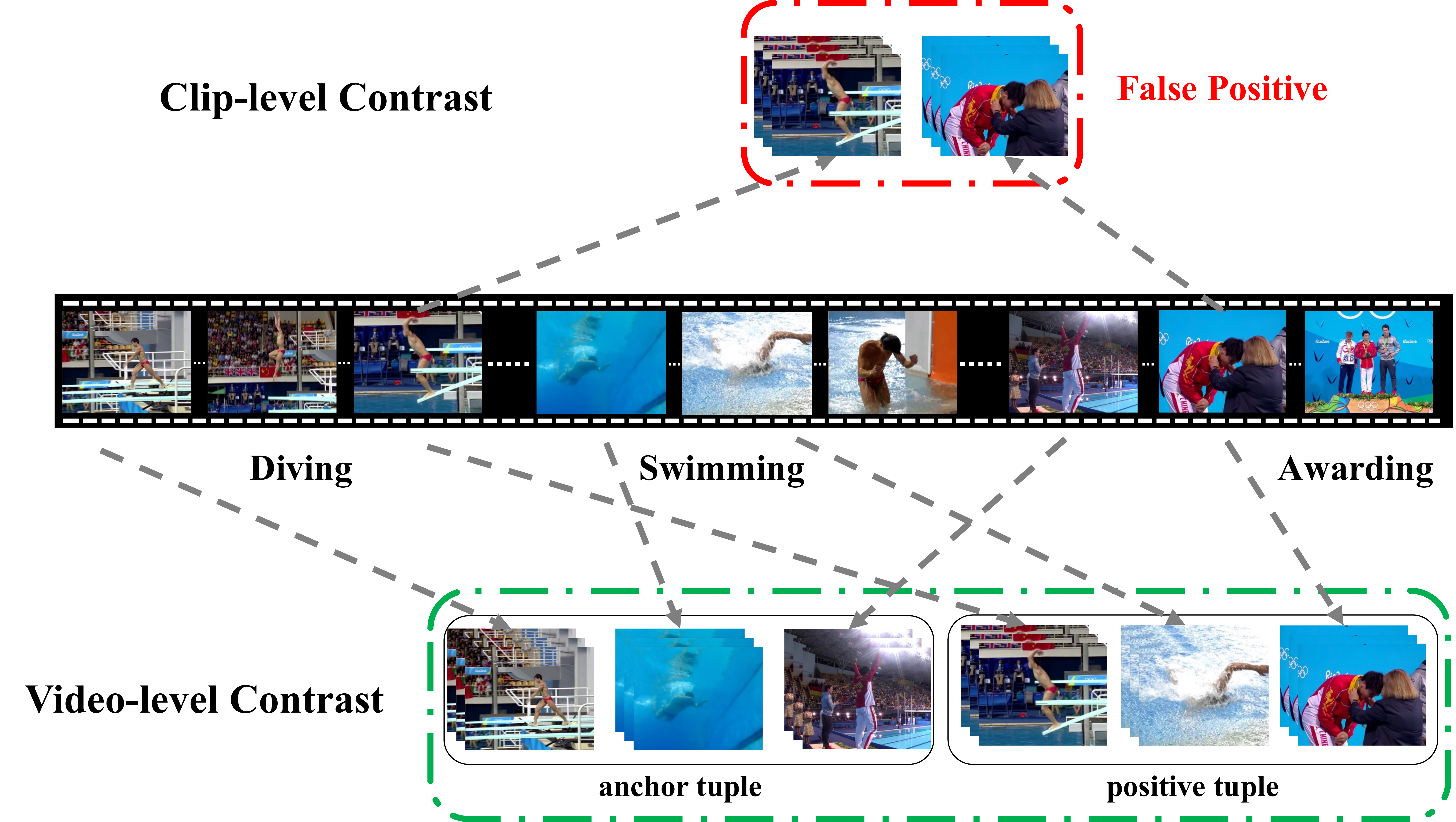}
		\end{center}
		\vspace{-2ex}
		\caption{Random sampling of clips from a long video may generate false positive pairs. Our proposed video-level contrastive framework based on segments can generate robust positive training pairs and reason the global information.}
		\label{fig:teaser}
		\vspace{-2ex}
	\end{figure}
	
	A key component of contrastive learning is to define positive and negative samples to contrast. For example, in the image domain, most approaches use random crops from the same image as a positive pair and consider other images in the dataset as negative samples \cite{wu_cvpr2018_instdisc,he_cvpr2020_moco,chen_icml2020_simclr}. For video, there are several widely adopted formulations. (i) Instance-based methods \cite{gordon_arxiv2020_vince,yao_aaai2021_seco,qian_cvpr2021_cvrl,han_nips2020_coclr} use random frames/clips from the same video as a positive pair.
	(ii) Pace-based methods \cite{wang_eccv2020_pace,yang_arxiv2020_vthcl} use clips from the same video but with different sampling rates as a positive pair.
	(iii) Prediction-based methods \cite{han_iccvw2019_dpc,han_eccv2020_memdpc} use predictions from auto-regressed and ground-truth features at the same spatiotemporal location as a positive pair. However, all previous methods define positive pairs to perform contrastive learning on frame-level or clip-level, which do not capture the global context of a video in long temporal range. In addition, random sampling of clips from an untrimmed video may generate false positive pairs as shown in Fig. \ref{fig:teaser} top.
	
	In this paper, we propose a new formulation for generating positive pairs to address the above limitation. As shown in Fig. \ref{fig:teaser} bottom, we first uniformly divide the video into several segments, and randomly pick a clip from each segment to form the anchor tuple. Then we randomly pick a clip from each segment again to form the positive tuple.  We consider these two tuples as a positive pair, and tuples from other videos as negative samples. 
	In this way, our formulation of positive samples is flexible (i.e., can be trained on any videos regardless of duration or sampling rate) and is able to reason about the global information across a video. However, global context alone is not enough to guide the unsupervised video representation learning since it does not enforce the inherent sequential structure of a video. Hence, we introduce a regularization loss based on the temporal order constraint. To be specific, we shuffle the frame order inside each tuple and ask the model to predict if the tuple has the correct temporal order. Since both proposed techniques work on the video-level, we term our method as \textit{VCLR, video-level contrastive learning}. Our contributions can be summarized as follows.
	
	\begin{itemize}
		\item We introduce a new way to formulate positive pairs based on segments for video-level contrastive learning.
		\item We incorporate a temporal order regularization term to enforce the inherent sequential structure of videos.
		\item Our proposed video-level contrastive learning framework (VCLR) outperforms previous literature on five datasets for downstream action classification, action localization and video retrieval.
	\end{itemize}

	\section{Related Work}
	\label{sec:related_work}
	\noindent \textbf{Self-supervised image representation learning}
	Initial attempts for self-supervised image representation learning are via designing various pretext tasks. Recently, a type of contrastive learning method based on instance discrimination has taken-off as it has been consistently demonstrated to outperform its supervised counterparts on downstream tasks \cite{chen_arxiv2020_mocov2,chen_arxiv2020_simclrv2,grill_arxiv2020_byol,caron_nips2020_swav}. The core idea behind contrastive learning is to learn representations by distinguishing between similar and dissimilar instances \cite{wu_cvpr2018_instdisc}. More specifically, it obtains the supervision signal by designing different forms of positive and negative pairs, such as random crops from the same image \cite{he_cvpr2020_moco,chen_icml2020_simclr}, different views of the same instance \cite{tian_eccv2020_cmc}, etc. 
	
	\noindent \textbf{Self-supervised video representation learning}
	Compared with images, videos have another axis, temporal dimension, which we can use to craft pretext tasks for self-supervised representation learning. 
	There are many previous attempts~\cite{zhu_arxiv2020_survey} including predicting the future \cite{vondrick_icml2015_lstm,vondrick_cvpr2016_anticipating,luo_cvpr2017_unsupervised,diba_iccv2019_dynamoNet}, predicting the correct order of shuffled frames \cite{misra_eccv2016_shuffle, fernando_cvpr2017_o3n} or video clips \cite{lee_iccv2017_opn,xu_cvpr2019_vcop}, predicting video rotation \cite{jing_arxiv2018_videoRotation}, solving a space-time cubic puzzle \cite{kim_aaai2019_cubicPuzzle}, predicting motion and appearance statistics \cite{wang_cvpr2019_statistics}, predicting speed \cite{benaim_cvpr2020_speednet,yao_cvpr2020_prp} and exploring video correspondence \cite{wei_cvpr2018_aot,wang_cvpr2019_cycleTime,dwibedi_cvpr2019_tcc,jabri_nips2020_walk}. There is another line of work using multi-modality signals as supervision, such as audio \cite{korbar_2018_avts,alayrac_nips2020_mmv,alwassel_2020_xdc,patrick_2020_gdt,piergiovanni_2020_elo} and language \cite{miech_2020_milnce,alayrac_nips2020_mmv}.
	
	Inspired by the great success of using contrastive learning in image domain, several recent work have considered it in video domain. Despite their loss objectives are similar (e.g., NCE \cite{oord_arxiv2018_cpc} and its variant), their major differences lie in how they formulate positive and negative samples. 
	CBT \cite{sun_arxiv2019_cbt} adapts masked language modeling \cite{devlin_arxiv2018_bert} to a video sequence, and uses the video clip and its masked version as a positive pair.
	DPC \cite{han_iccvw2019_dpc} and MemDPC \cite{han_eccv2020_memdpc} use predictions from autoregressed and ground-truth features at the same spatiotemporal location as a positive pair. 
	Pace \cite{wang_eccv2020_pace} and VTHCL \cite{yang_arxiv2020_vthcl} proposes to use video clips of the same action instance but with different visual tempos as a positive pair.  
	CMC \cite{tian_eccv2020_cmc} introduces two ways of sampling positive pairs: (1) different frames from the same video and (2) RGB and flow data of the same frame. 
	Similar to CMC, VINCE \cite{gordon_arxiv2020_vince} and SeCo \cite{yao_aaai2021_seco} also use different frames from the same video, while CVRL \cite{qian_cvpr2021_cvrl} use different clips from the same video as positive samples. There are also many concurrent work~\cite{wang_aaai2021_dsm, chen_aaai2021_rspnet, pan_cvpr2021_videomoco, wang_cvpr2021_be, feichtenhofer_cvpr2021_large_spatiotemporal, guo_cvpr2021_ssan, huang_cvpr2021_cmd, wang_cvpr2021_ctp}.
	
	Different from previous approaches, our work proposes a new way to formulate positive pairs based on segments.
	Given our formulation, we can perform contrastive learning on video-level rather than frame- or clip-level, and achieve promising results both quantitatively and qualitatively.
	In addition, our method is generally applicable to various network architectures, and orthogonal to other advancements in self-supervised video representation learning domain.

	\section{Methodology}
	\label{sec:methodology}
	
	\begin{figure*}[t]
		\begin{center}
			\includegraphics[width=2.0 \columnwidth]{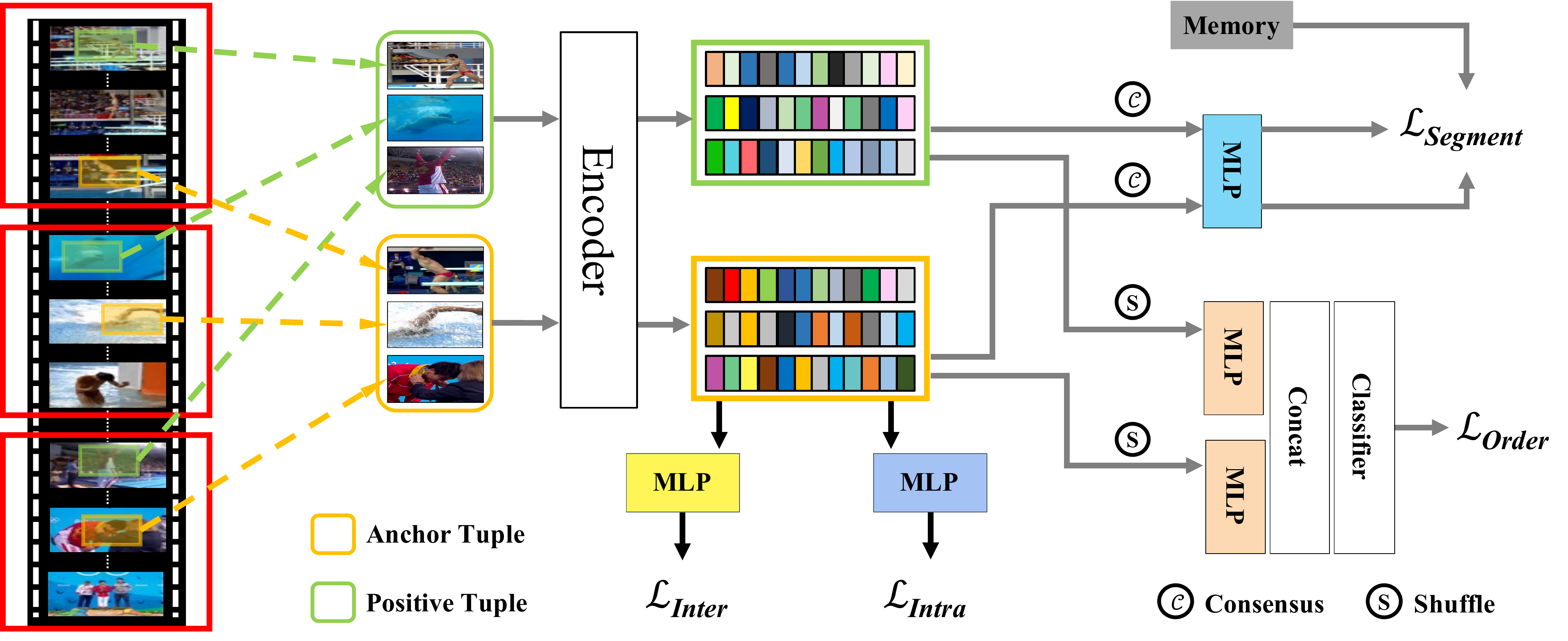}
		\end{center}
		\vspace{-2ex}
		\caption{\textbf{Overview of the proposed VCLR framework}. We introduce a new way to formulate positive training pairs based on video segments. There are four loss objectives, $\mathcal{L}_{\textit{Inter}}$ and $\mathcal{L}_{\textit{intra}}$ following \cite{yao_aaai2021_seco}, and two new ones $\mathcal{L}_{\textit{Segment}}$ and $\mathcal{L}_{\textit{Order}}$ for video-level contrastive learning. Image is best viewed in color.}
		\label{fig:overview}
		\vspace{-2ex}
	\end{figure*}
	
	\subsection{Preliminary}
	\label{subsec:preliminary}
	Before diving into our proposed framework, we first revisit the concept of contrastive learning in image domain \cite{wu_cvpr2018_instdisc,he_cvpr2020_moco,chen_icml2020_simclr}. 
	Here we use MoCoV2 \cite{chen_arxiv2020_mocov2} as an illustrating example. More formally, given a set of images $\mathcal{X}$, an image $x_{i}$ is sampled from $\mathcal{X}$ and is augmented to generate a positive
	pair $x^{a}_{i}$ and $x^{+}_{i}$.
	A set of negative samples $\mathcal{N^{-}}$ are then selected from the rest of $\mathcal{X}$, i.e., $x_{j} \in \mathcal{N^{-}}, j \neq i$. Two encoders, query encoder $f_{q}$ and key encoder $f_{k}$, are used to obtain the visual representations, e.g., 2048-dim features from ResNet50 backbone. These visual representations are then projected via MLP heads $g_{q}$ and $g_{k}$ to lower dimensional embeddings for similarity comparison. 
	For notation simplicity, we represent anchor embedding as $q = g_{q}(f_{q}(x^{a}_{i}))$, positive embedding as $p = g_{k}(f_{k}(x^{+}_{i}))$, and negative embeddings as $n_{j} = g_{k}(f_{k}(x^{-}_{j}))$.
	At this point, the problem is formulated as a (N+1)-way classification task, and can be optimized by InfoNCE loss \cite{oord_arxiv2018_cpc} as
	\begin{equation}
	\mathcal{L}_{\text{NCE}}(q, p, \mathcal{N^{-}}) = -\log \frac{e^{\text{sim}(q, p)}}{e^{\text{sim}(q, p)} + \sum_{j = 1}^{N} e^{\text{sim}(q, n_{j})}}.
	\label{eq:infonce}
	\end{equation}
	Here, sim($\cdot$) is the distance metric used to measure the similarity between feature embeddings, e.g., dot product. $N$ denotes the number of negative samples.
	By optimizing the objective, the model learns to map similar instances closer and dissimilar instances farther apart in the embedding space. For more details, we refer the readers to \cite{he_cvpr2020_moco,chen_arxiv2020_mocov2}.
	
	To adapt contrastive learning into video domain, we need to find a way to define positive and negative samples. We have mentioned several in Sec. \ref{sec:related_work}. In this paper, we follow a recent state-of-the-art SeCo \cite{yao_aaai2021_seco} as our baseline given their strong performance.
	Specifically, they introduce two ways to formulate positive and negative samples: inter-frame and intra-frame instance discrimination. 
	
	\noindent \textbf{Inter-frame instance discrimination} Unlike approaches in image domain that take random crops from the same image as a positive pair, inter-frame instance discrimination simply treats random frames from the same video as positive. 
	Suppose we randomly pick three frames from a video, $v_{1}, v_{2}$ and $v_{3}$, inter-frame instance discrimination task considers $v^{a}_{1}$ as the anchor frame, $(v^{+}_{1}, v_{2}, v_{3})$ as the positive samples, and frames from other videos in the dataset as negative samples $\mathcal{N^{-}}$. Here, $v^{a}_{1}$ and $v^{+}_{1}$ are generated from $v_{1}$ by different augmentations. 
	Mathematically, the loss objective can be written as  
	\begin{equation}
	\mathcal{L}_{\textit{Inter}} = \frac{1}{3}\sum \mathcal{L}_{\text{NCE}}(q_{1}^{a}, \{p_{1}^{+}, p_{2}, p_{3}\}, \mathcal{N^{-}}).
	\label{eq:inter_frame}
	\end{equation}
	We represent anchor embeddinig as $q_{1}^{a} = g_{q}^{e}(f_{q}(v^{a}_{1}))$, and positive embeddings as $p_{1}^{+} = g_{k}^{e}(f_{k}(v^{+}_{1}))$, $p_{2} = g_{k}^e{}(f_{k}(v_{2}))$, $p_{3} = g_{k}^{e}(f_{k}(v_{3}))$. 
	$g_{q}^{e}$ and $g_{k}^{e}$ are MLP heads for inter-frame instance discrimination task.
	Note that, we have multiple positives in Eq. (\ref{eq:inter_frame}), we compute the contrastive losses separately and take their average. 
	
	\noindent \textbf{Intra-frame instance discrimination} 
	In order to learn better appearance representation, intra-frame instance discrimination is adopted to explore the inherently spatial changes across frames. 
	Different from inter-frame, we only use $v^{+}_{1}$ as a positive sample, and consider $v_{2}$, $v_{3}$ as negative samples. Hence, the loss objective can be written as 
	\begin{equation}
	\mathcal{L}_{\textit{Intra}} = \mathcal{L}_{\text{NCE}}(q^{a}_{1}, \{p_{1}^{+}\}, \{p_{2}, p_{3}\}).
	\label{eq:intra_frame}
	\end{equation}
	Essentially, this task is the same as standard instance discrimination in image domain except trained with a much smaller negative set. 
	The embeddings in Eq. (\ref{eq:intra_frame}) are actually different from the ones in Eq. (\ref{eq:inter_frame}) due to using different MLP heads $g_{q}^{r}$ and $g_{k}^{r}$ for intra-frame instance discrimination task, but we reuse them for notation simplicity.
	
	Despite the design of both inter- and intra- frame instance discrimination task, SeCo still performs contrastive learning on the frame-level. Simply put, intra-frame is between two crops in the same image, and inter-frame is between two frames in the same video.
	Some recent work \cite{han_nips2020_coclr,qian_cvpr2021_cvrl} use short video clips as input to perform instance discrimination, which can be considered doing contrastive learning in the clip-level.
	However, contrastive learning on frame- or clip-level are sub-optimal for video representation learning because they cannot capture the evolving semantics in the temporal dimension.
	Hence, we ask the question, how can we perform contrastive learning in the video-level?

	\subsection{Video-level contrastive learning}
	\label{subsec:vclr}
	Long-range temporal structure plays an important role in understanding the dynamics in videos. 
	In terms of supervised learning, there have been a series of work to explore global context \cite{wang_eccv2016_tsn,diba_cvpr2017_tle,girdhar_cvpr2017_actionVLAD,wang_cvpr2018_nonlocal,zhu_accv2018_hidden,wu_cvpr2019_featureBank,zhang_iclr2020_V4D,li_arxiv2020_nuta,li_iccv2021_vidtr}. 
	However, in terms of self-supervised learning, few efforts have been done to incorporate global video-level information.
	
	In this work, inspired by \cite{wang_eccv2016_tsn,lan_cvprw2017_dvof}, we propose to use the segment idea to sample positive pairs for contrastive learning. 
	Formally, given a video $V$, we divide it into $K$ segments $\{S_{1}, S_{2}, \cdots, S_{K}\}$ of the equal duration. Within each segment $S_{k}$, we randomly sample a frame 
	$v_{k}$ to formulate our anchor tuple, i.e., $t^{a} = \{v_{1}^{a}, v_{2}^{a}, \cdots, v_{K}^{a}\}$. Then, we take a second independent random sample of $K$ frames in the same manner to formulate the positive tuple, $t^{+} = \{v_{1}^{+}, v_{2}^{+}, \cdots, v_{K}^{+}\}$. 
	We consider these two tuples as a positive pair, because both of them describe the evolving semantics inside a video.
	Intuitively, segment-based sampling can be considered as a form of data augmentation, so that the consensus from  two tuples can produce two different views of the same video, which is what instance discrimination need to learn an effective representation.
	
	In terms of loss objective, each frame in the tuple will produce its own preliminary frame-level representation. Then a consensus among these representations will be derived as the video-level representation. The final embedding of the anchor and positive tuple can be represented as 
	\begin{equation}
	q_{t}^{a} = g_{q}^{s} (\mathcal{C} [f_{q}(v_{1}^{a}), f_{q}(v_{2}^{a}), ..., f_{q}(v_{K}^{a})]) 
	\label{eq:anchortuple}
	\end{equation}
	\begin{equation}
	p_{t}^{+} = g_{k}^{s} (\mathcal{C} [f_{k}(v_{1}^{+}), f_{k}(v_{2}^{+}), ..., f_{k}(v_{K}^{+})])
	\label{eq:positivetuple}
	\end{equation}
	where $\mathcal{C}$ denotes the consensus operation, e.g., average.
	$g_{q}^{s}$ and $g_{k}^{s}$ are MLP heads for video-level instance discrimination task.
	Once we have the embeddings $q_{t}^{a}$ and $p_{t}^{+}$, we can compute the video-level contrastive loss as
	\begin{equation}
	\mathcal{L}_{\textit{segment}} = \mathcal{L}_{\text{NCE}}(q_{t}^{a}, p_{t}^{+}, \mathcal{N^{-}}).
	\label{eq:vclr}
	\end{equation}
	Here, we reuse the notation $\mathcal{N^{-}}$ to generally indicate the negative set, where each negative is a tuple of frames sampled from other videos in the dataset.
	
	We want to point it out that, this formulation of positive samples has several advantages. First, it leads to a more robust training as it sees information from the entire video. Second, we can use this strategy to train a model on any types of video, regardless of duration or sampling rate. Third, this formulation is not limited to video frames with 2D CNNs, but also ready to be used for video clips with 3D CNNs. 
	In addition, our formulation is orthogonal to other advancements in self-supervised video representation learning \cite{yang_arxiv2020_vthcl,han_nips2020_coclr,han_iccvw2019_dpc}, which could be incorporated seamlessly.
	
	\begin{figure}[t]
		\captionsetup[subfigure]{labelformat=empty}
		\centering
		\subfloat[]{
			\includegraphics[width=0.48  \columnwidth]{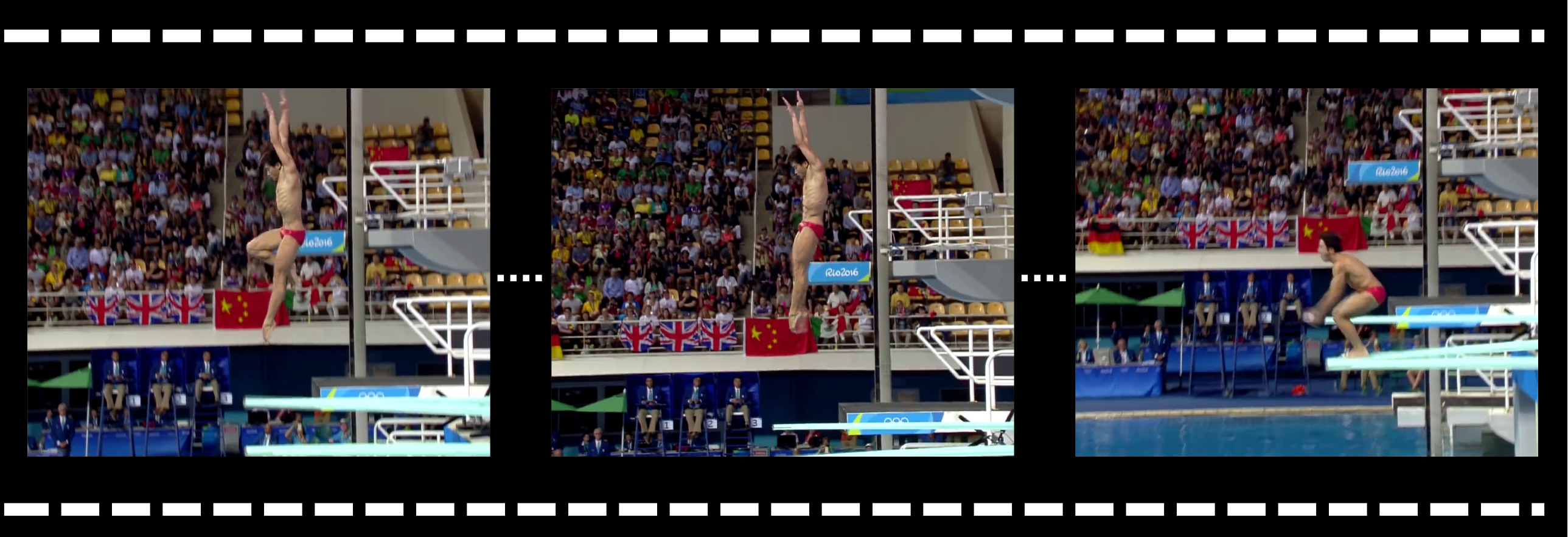}
		}
		\subfloat[]{
			\includegraphics[width=0.48  \columnwidth]{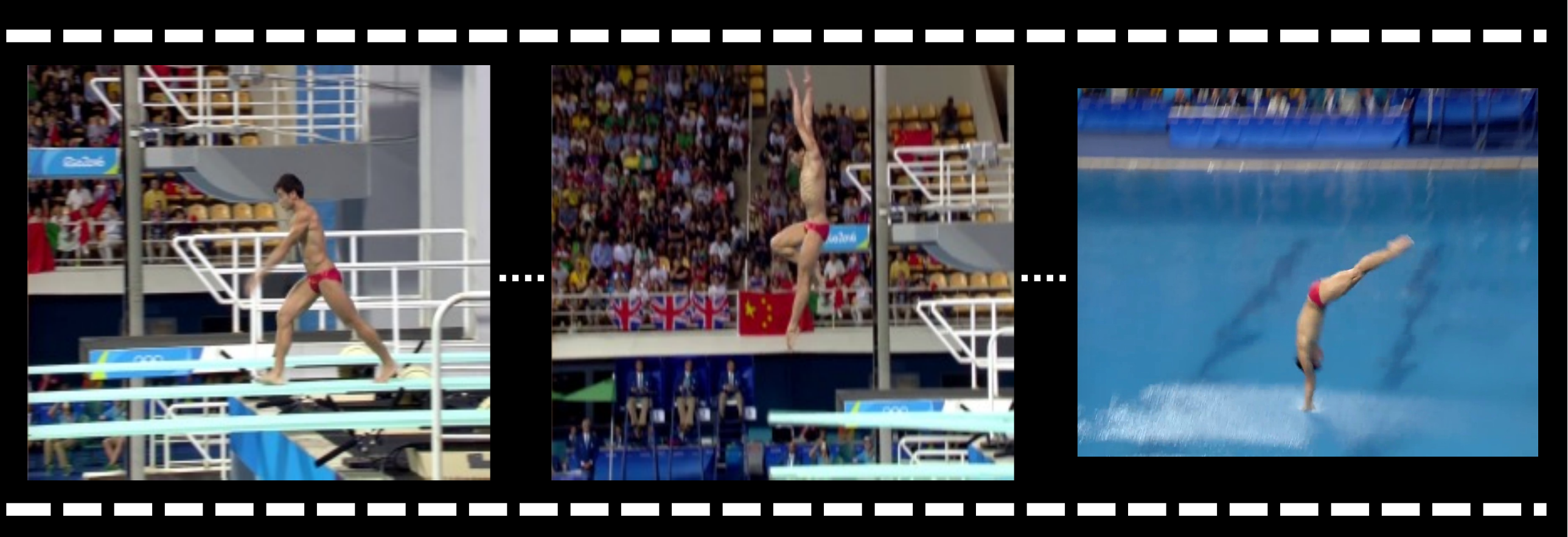}
		}
		\vspace{-3ex}
		\caption{\textbf{Left}: It is challenging to identify the temporal order within a local sequence, which will likely confuse the model training. \textbf{Right}: With longer temporal context, we can easily tell the evolution of events.}
		\label{fig:order_sample}
		\vspace{-2ex}
	\end{figure}

	\begin{table}[t]
		\centering
		\scalebox{0.8}{
			\begin{tabular}{l|c|c}
				\hline
				Method            & Network  & Top-1 Acc. (\%) \\ 
				\hline
				VTHCL \cite{yang_arxiv2020_vthcl}             & R3D-50                     & 37.8            \\
				VINCE  \cite{gordon_arxiv2020_vince}           & R2D-50                         & 49.1            \\
				SeCo  \cite{yao_aaai2021_seco}            & R2D-50                              & 61.9            \\ \hline
				VCLR              & R2D-50                           & \textbf{64.1}           \\ \hline
				\textcolor{gray}{Sup-ImageNet} & \textcolor{gray}{R2D-50}                          & \textcolor{gray}{52.3}           \\
				\textcolor{gray}{Sup-Kinetics400} & \textcolor{gray}{R2D-50}                          & \textcolor{gray}{69.9}           \\ \hline
			\end{tabular}
		}
		\caption{\textbf{Linear evaluation on Kinetic400}. Our method outperforms previous methods and is close to supervised upper bound. Sup-ImageNet:supervised ImageNet weights. Sup-Kinetics400: supervised training on Kinetics400.}
		\label{tab:k400}
		\vspace{-2ex}
	\end{table}
	
	\subsection{Temporal order regularization}
	\label{subsec:temporal}
	Video-level contrastive learning helps to capture global context as it sees information from the entire video, but it is weak in enforcing the inherent sequential structure. 
	Given temporal coherence is a strong constraint, there has been some work \cite{misra_eccv2016_shuffle,fernando_cvpr2017_o3n,lee_iccv2017_opn,yao_aaai2021_seco} that use it as a supervision signal for self-supervised video representation learning.
	
	In this paper, we also use the temporal order as a regularization term under our contrastive learning framework. To be specific, given tuples $t^{a}$ and $t^{+}$, which are sampled based on segments described in Sec. \ref{subsec:vclr}, we randomly shuffle the frames inside each tuple with $50\%$ chance. This will lead to a balanced 4-way classification problem: both tuples are in correct temporal order, $t^{a}$ correct and $t^{+}$ shuffled, $t^{a}$ shuffled and $t^{+}$ correct, and both shuffled.
	
	In terms of loss objective, we first compute the embeddings of each tuple as $o_{t}^{a}$ and $o_{t}^{+}$,
	\begin{equation}
	o_{t}^{a} = [g_{q}^{o} (f_{q}(v_{1}^{a})), g_{q}^{o} (f_{q}(v_{2}^{a})), ..., g_{q}^{o} (f_{q}(v_{K}^{a}))]
	\label{eq:anchortuple_order}
	\end{equation}
	\begin{equation}
	o_{t}^{+} = [g_{k}^{o} (f_{k}(v_{1}^{+})), g_{k}^{o} (f_{k}(v_{2}^{+})), ..., g_{k}^{o} (f_{k}(v_{K}^{+}))]
	\label{eq:positivetuple_order}
	\end{equation}
	$g_{q}^{o}$ and $g_{k}^{o}$ are MLP heads for temporal order regularization.
	In order to keep the temporal order within each tuple, we concatenate the embeddings and predict its order type,
	\begin{equation}
	y^{*} = h_o([\text{concat}(o_{t}^{a}), \text{concat}(o_{t}^{+})]).
	\label{eq:order_prediction}
	\end{equation}
	Here, $h_{o}$ is a linear classifier to project the concatenated embeddings to logits. $y^{*}$ is the order type prediction, i.e., class 0, 1, 2 or 3. In the end, we compute the cross-entropy loss between the predictions and pre-defined ground-truth 
	\begin{equation}
	\mathcal{L}_{\textit{Order}} (y, y^{*})= -\sum y \log y^{*}.
	\label{eq:temporal_loss}
	\end{equation}
	
	Our temporal order regularization may seem similar to \cite{fernando_cvpr2017_o3n,lee_iccv2017_opn}, but we build it upon our video-level sampling framework. Simply put, previous approaches use local sequence to validate temporal order, while we use global sequence. One drawback of using local sequence is that it may not provide enough cues to predict the correct sequential structure. 
	As shown in Fig.~\ref{fig:order_sample} left, any temporal order seems reasonable between these three frames sampled from a local sequence. 
	On the contrary in Fig.~\ref{fig:order_sample} right, using our video-level sampling strategy, it is less ambiguous to determine the temporal order inside a tuple. 
	
	\subsection{Overall framework}
	\label{subsec:overall}
	At this point, we present our video-level contrastive learning method, VCLR, for self-supervised video representation learning. Our overall framework can be seen in Fig. \ref{fig:overview}. During training, the final loss objective is a summation of previous four loss objectives,
	\begin{equation}
	\mathcal{L} = \mathcal{L}_{\textit{Inter}} + \mathcal{L}_{\textit{Intra}} + \mathcal{L}_{\textit{Segment}} + \mathcal{L}_{\textit{Order}}
	\label{eq:overall_loss}
	\end{equation}
	For downstream tasks, we ignore the MLP heads designed for each task, e.g., $g^{e}$, $g^{r}$, $g^{s}$ and $g^{o}$. We only use pretrained encoder $f_{q}$ for extracting features or finetuning.

	\section{Experiments}
	\label{sec:experiments}

	\subsection{Datasets}
	\label{subsec:datasets}
	We conduct experiments on 5 datasets, Kinetics400 \cite{kinetics400}, UCF101 \cite{ucf101}, HMDB51 \cite{hmdb51}, Something-Something-v2 \cite{sthsth} and ActivityNet \cite{heilbron_cvpr2015_activitynet}. 
	\textbf{Kinetics400} consists of approximately 240k training and 20k validation videos trimmed to 10 seconds from 400 human action categories. 
	\textbf{UCF101} contains $13,320$ videos spreading over 101 categories of human actions.
	\textbf{HMDB51} contains $6,849$ videos divided into 51 action categories.
	\textbf{Something-Something-v2} consists of 174 action classes and a total of $220,847$ videos. For notation simplicity, we refer this dataset as SthSthv2 for the rest of the paper. 
	\textbf{ActivityNet} (V1.3) contains 200 human daily living actions. It has $10,024$ training and $4,926$ validation videos.
	Both UCF101 and HMDB51 have three official train-val splits, and we report performance on its split 1 for fair comparison to previous work.
	For other three datasets, we report performance on their validation sets.
	
	\begin{table}[t]
		\centering
		\scalebox{0.8}{
			\begin{tabular}{l|c|c|cc}
				\hline
				\multirow{2}{*}{Method} & \multirow{2}{*}{\begin{tabular}[c]{@{}c@{}}Venue\end{tabular}} &  
				\multirow{2}{*}{Network}
				&
				\multicolumn{2}{c}{Top-1 Acc (\%)} \\
				&    &  & UCF101  & HMDB51        \\ 
				\hline
				ST-Puzzle \cite{kim_aaai2019_cubicPuzzle}              & AAAI19   & R3D-50 & 65.8                  & 33.7 \\ 
				MAS  \cite{wang_cvpr2019_statistics}            & CVPR19    & C3D & 61.2         & 33.4        \\
				DPC  \cite{han_iccvw2019_dpc}                   & ICCVW19   & R-2D3D & 75.7                  & 35.7        \\
				SpeedNet  \cite{benaim_cvpr2020_speednet}    & CVPR20    & S3D-G   & 81.1                  & 48.8        \\
				VIE  \cite{zhuang_cvpr2020_VIE}    & CVPR20   & SlowFast & 80.4                  & 52.5        \\
				MemDPC  \cite{han_eccv2020_memdpc}      & ECCV20   & R-2D3D  & 78.1                  & 41.2        \\
				PacePred  \cite{wang_eccv2020_pace}              & ECCV20  & R(2+1)D  & 77.1                  & 36.6        \\
				TT  \cite{jenni_eccv2020_temporalssl}              & ECCV20  & R3D-18  & 79.3                  & 49.8        \\
				SeCo\footnotemark \cite{yao_aaai2021_seco}    & AAAI21   & R2D-50 & 83.4                      & 49.7       \\
				\hline
				VCLR                    & -           & R2D-50 & \textbf{85.6}        & \textbf{54.1}       \\ 
				\hline
				\textcolor{gray}{Sup-ImageNet}              & -    & \textcolor{gray}{R2D-50}                               & \textcolor{gray}{81.6}                  & \textcolor{gray}{49.0}       \\ 
				\textcolor{gray}{Sup-Kinetics400}              & -    & \textcolor{gray}{R2D-50}                                & \textcolor{gray}{88.1}                  & \textcolor{gray}{56.1}       \\ \hline
			\end{tabular}
		}
		\caption{\textbf{Downstream action classification on UCF101 and HMDB51}.  Sup-Kinetics400: supervised Kinetics400 pretrained weights.}
		\label{tab:ucf_hmdb51_finetune}
		\vspace{-2ex}
	\end{table}
	
	\footnotetext[1]{SeCo \cite{yao_aaai2021_seco} reported higher numbers on UCF101 and HMDB51. We follow the default hyperparameters from MemDPC \cite{han_eccv2020_memdpc} to perform  finetuning on all R2D-50 based methods, so the comparisons in Table 2 are fair.}
	
	\begin{table}[t]
		\centering
		\scalebox{0.8}{
			\begin{tabular}{l|c|c}
				\hline
				Method    & Video Algorithm   & Top-1 Acc (\%) \\ 
				\hline
				\multirow{2}{*}{VINCE \cite{gordon_arxiv2020_vince}}   & TSN       & 31.4   \\
				& TSM      & 50.3 \\ \hline
				\multirow{2}{*}{SeCo \cite{yao_aaai2021_seco}}   & TSN       & 31.9   \\
				& TSM     & 50.7 \\ \hline
				\multirow{2}{*}{VCLR}   & TSN       & \textbf{33.3}   \\
				& TSM      & \textbf{52.0} \\ \hline
				\multirow{2}{*}{\textcolor{gray}{Sup-ImageNet}} & \multicolumn{1}{c|}{\textcolor{gray}{TSN}}             & \textcolor{gray}{33.0}          \\
				& \multicolumn{1}{c|}{\textcolor{gray}{TSM}}   & \textcolor{gray}{59.1}          \\ \hline
			\end{tabular}
		}
		\caption{\textbf{Downstream action classification on SthSthv2}. TSM: temporal shift module network \cite{lin_iccv2019_tsm}. All methods in this table use R2D-50 as encoder, and follow the same training setting for fair comparison.}
		\label{tab:sthsthv2_finetune}
		\vspace{-2ex}
	\end{table}

	\begin{table*}[t]
		\centering
		\scalebox{0.8}{
			\begin{tabular}{lccc|cccc|cccc}
				\hline
				\multicolumn{1}{l}{\multirow{2}{*}{Method}} & \multicolumn{1}{c}{\multirow{2}{*}{Modality}} & \multicolumn{1}{c}{\multirow{2}{*}{Network}} & \multirow{2}{*}{Pretrain} & \multicolumn{4}{c|}{UCF101}                       & \multicolumn{4}{c}{HMDB51} \\
				\multicolumn{1}{c}{}                        & \multicolumn{1}{c}{}                      &    &                      & R@1  & R@5  & R@10                      & R@20 & R@1  & R@5  & R@10 & R@20 \\ \hline
				OPN \cite{lee_iccv2017_opn}  & V   &  VGG                                   & UCF101                      & 19.9 & 28.7 & 34.0                      & 40.6 & -    & -    & -    & -    \\
				DRL \cite{buchler_eccv2018_drl} & V  & VGG                                     & UCF101                      & 25.7 & 36.2 & 42.2                      & 49.2 & -    & -    & -    & -    \\
				VCOP  \cite{xu_cvpr2019_vcop} & V  & R(2+1)D                                     & UCF101                      & 14.1 & 30.3 & 40.4                      & 51.1 & 7.6  & 22.9 & 34.4 & 48.8 \\
				VCP  \cite{luo_aaai2020_vcp}    & V &  R3D-50                                     & UCF101                      & 18.6 & 33.6 & 42.5                      & 53.5 & 7.6  & 24.4 & 36.3 & 53.6 \\
				MemDPC  \cite{han_eccv2020_memdpc}   & V &    R-2D3D                                    & UCF101                      & 20.2 & 40.4 & 52.4                      & 64.7 & 7.7  & 25.7 & 40.6 & 57.7 \\
				\textcolor{gray}{MemDPC \cite{han_eccv2020_memdpc}}   & \textcolor{gray}{F} &  \textcolor{gray}{R-2D3D} & \textcolor{gray}{UCF101}                      & \textcolor{gray}{40.2} & \textcolor{gray}{63.2} & \textcolor{gray}{71.9}                      & \textcolor{gray}{78.6} & \textcolor{gray}{15.6} & \textcolor{gray}{37.6} & \textcolor{gray}{52.0} & \textcolor{gray}{65.3} \\
				\textcolor{gray}{CoCLR \cite{han_nips2020_coclr}}   & \textcolor{gray}{VF} &  \textcolor{gray}{S3D} & \textcolor{gray}{UCF101}                      & \textcolor{gray}{53.3} & \textcolor{gray}{69.4} & \textcolor{gray}{76.6}                      & \textcolor{gray}{82.0} & \textcolor{gray}{23.2} & \textcolor{gray}{43.2} & \textcolor{gray}{53.5} & \textcolor{gray}{65.5} \\
				VCLR   & V & R2D-50 & UCF101      & \textbf{46.8} & \textbf{61.8} & \textbf{70.4}                      & \textbf{79} & \textbf{17.6}  & \textbf{38.6} & \textbf{51.1} & \textbf{67.6} \\
				\hline
				SpeedNet \cite{benaim_cvpr2020_speednet}     & V           &     S3D-G                        & Kinetics400                     & 13.0 & 28.1 & 37.5                      & 49.5 & -    & -    & -    & -    \\
				SeCo  \cite{yao_aaai2021_seco}   & V  & R2D-50                                     & Kinetics400                     & 69.5 & 79.5 & 85                        & 90.1 & 33.6 & 57.4 & 67.6 & 78.6 \\
				VCLR        &   V   &  R2D-50  & Kinetics400                     & \textbf{70.6} & \textbf{80.1} & \multicolumn{1}{c}{\textbf{86.3}} & \textbf{90.7} & \textbf{35.2} & \textbf{58.4} & \textbf{68.8} & \textbf{79.8} \\ \hline
			\end{tabular}
		}
		\caption{\textbf{Downstream video retrieval on UCF101 and HMDB51}. V: RGB frames. F: optical flow. }
		\label{tab:ucf_hmdb_retrieval}
		\vspace{-2ex}
	\end{table*}
	
	\begin{table}[t]
		\centering
		\scalebox{0.8}{
			\begin{tabular}{l|ccc}
				\hline
				\multirow{2}{*}{Method}     & Classification & \multicolumn{2}{c}{Localization} \\
				& Top-1 Acc (\%) & AUC (\%)      & AR@100 (\%)      \\ \hline
				
				VINCE \cite{gordon_arxiv2020_vince}      & 60.7      & 64.6     & 73.2     \\ 
				SeCo \cite{yao_aaai2021_seco}       & 67.8         & 65.2     & 73.4    \\  
				VCLR        & \textbf{71.9}     & \textbf{65.5}     & \textbf{73.8}     \\
				\hline
				\textcolor{gray}{Sup-ImageNet}    & \textcolor{gray}{67.2}      & \textcolor{gray}{64.8}     & \textcolor{gray}{73.4}     \\ 
			\end{tabular}
		}
		\caption{\textbf{Downstream action classification and localization on ActivityNet}. All methods use R2D-50 as encoder, and follow the same training setting for fair comparison.}
		\label{tab:activitynet_finetune}
		\vspace{-2ex}
	\end{table}

	\subsection{Implementation details of pretraining}
	\label{subsec:implementation}
	In terms of input, we follow our segment-based sampling strategy to choose two tuples from the same video as video-level anchor and positive sample. We set the number of segments to $3$, i.e., each tuple contains three frames. The video-level anchor will also be used to generate frame-level anchor and positive sample to compute the inter-frame and intra-frame losses. 
	For data augmentation, we apply random scales, color-jitter, random grayscale, random Gaussian blur, and random horizontal flip.
	In terms of network architecture, we use ResNet50 \cite{he_cvpr2016_resnet} as our backbone, plus four MLP heads designed for each loss. 
	We use average operation for $\mathcal{C}$. 
	In terms of optimization, we train the model on Kinetics400 dataset with a batch size of $512$ for $400$ epochs following \cite{yao_aaai2021_seco}.
	We use SGD to optimize the network with an initial learning rate of $0.2$, and annealed to zero with a cosine decay scheduler. More details can be found in the Appendix Sec~\ref{sec:supp_implement}.
	
	\subsection{Linear evaluation on Kinetics400}
	\label{subsec:k400}
	In order to quantify the quality of learned representation, the most straightforward way is to treat the pretrained model as a feature extractor and train a classifier on top of the features to see its generalization performance.  
	
	Following \cite{yao_aaai2021_seco}, we uniformly sample 30 frames from each video, resize each frame with short edge of 256 and center-crop it to $224 \times 224$. We forward each frame through the frozen ResNet50 backbone, get the frame-level features and average them into video-level feature. A linear SVM is then trained on the video-level features of the Kinetics400 training set, and finally evaluated on its validation set. 
	
	As we can see in Table \ref{tab:k400}, our method outperforms previous literature in terms of linear evaluation on Kinetics400 dataset. Especially for fair comparison when using 2D ResNet50, our VCLR pretrained model is $11.8\%$ higher than ImageNet pretrained weights, and $2.2\%$ higher than previous state-of-the-art SeCo pretrained weights \cite{yao_aaai2021_seco}. In addition, we further close the gap between unsupervised feature learning and supervised upper bound ($64.1\%$ vs $69.9\%$). This clearly demonstrates the effectiveness of performing contrastive learning on video-level.
	We note that a recent work CVRL \cite{qian_cvpr2021_cvrl} performs slightly better than us. However, we conjecture it is because CVRL has been trained with a heavier architecture (3D ResNet50 \cite{hara_cvpr2018_3DResNet} with 31.7M parameters) and twice our epochs (800). 
	
	\subsection{Downstream action classification}
	\label{subsec:downstream_action_classification}
	A main goal of unsupervised learning is to learn features
	that are transferable \cite{he_cvpr2020_moco}. In both computer vision and natural language processing, pretraining a network on large datasets and finetuning it on smaller datasets is the de facto way to achieve promising results on downstream tasks. 
	
	Following previous work \cite{han_eccv2020_memdpc,yao_aaai2021_seco}, we transfer our pretrained backbone and finetune all its layers on UCF101 and HMDB51 datasets. We compare our method to recent literature in Table \ref{tab:ucf_hmdb51_finetune}. Note that there are more related work, but we only list methods that are pretrained on Kinetics400 and using RGB frames as input for fair comparison. A complete comparison including models trained on larger datasets \cite{diba_iccv2019_dynamoNet} or multi-modality \cite{han_nips2020_coclr} can be found in the Appendix Table~\ref{tab:ucf_hmdb51_finetune_full}. As can be seen in Table \ref{tab:ucf_hmdb51_finetune}, our method with 2D ResNet50 is able to outperform those using more advanced 3D CNNs \cite{zhuang_cvpr2020_VIE,benaim_cvpr2020_speednet}. Compared to previous state-of-the-art SeCo \cite{yao_aaai2021_seco}, we find that video-level contrastive learning brings $2.2\%$ and $4.4\%$ absolute performance improvement on UCF101 and HMD51, respectively.
	Similarly, our VCLR pretrained models achieve higher accuracy than ImageNet pretrained weights on both datasets, and perform very close to Kinetics400 pretrained weights.

	In order to have a complete understanding of our learned representation's transferrability, we further evaluate VCLR on another two popular action classification datasets, SthSthv2 and ActivityNet. This is because UCF101 and HMDB51 have small domain gap to Kinetics400, given they are all scene-focused datasets, e.g., most actions can be predicted correctly by using object or background prior. SthSthv2 is a motion-focused dataset and requires strong temporal reasoning because most activities cannot be inferred based on spatial features alone (e.g. opening something, covering something with something). ActivityNet contains long and untrimmed videos, and thus requires strong global context modeling.
	
	In terms of SthSthv2, we compare our learned representations to those from VINCE and SeCo in Table \ref{tab:sthsthv2_finetune}. 
	As we can see, our learned representation achieves higher accuracy than both of them. We also show that our pretrained backbone is compatible with different video action recognition algorithms. Besides TSN \cite{wang_eccv2016_tsn}, we adapt our pretrained weights to initialize a popular algorithm TSM \cite{lin_iccv2019_tsm} on SthSthv2 and obtain promising results. 
	In terms of ActivityNet, we also compare our learned representations to those from VINCE and SeCo in Table \ref{tab:activitynet_finetune} and show significant performance improvements. 
	
	To summarize, by performing contrastive learning on video-level, our proposed VCLR is able to outperform previous state-of-the-art approaches on 4 downstream action classification datasets, including scene-focused datasets (UCF101 and HMDB51), motion-focused dataset (SthSthv2) and untrimmed video dataset (ActivityNet).

	\subsection{Downstream action temporal localization}
	\label{subsec:downstream_temporal_localization}
	We also evaluate VCLR on action temporal localization task with ActivityNet. The goal of this task is to generate high quality proposals to cover action instances with high recall and high temporal overlap.
	To be specific, we adopt the popular BMN \cite{lin_iccv2019_bmn} method for action localization, and only change the input features to see which representation generalizes better. To evaluate proposal quality, Average Recall (AR) under multiple IoU thresholds are calculated. We calculate AR under different Average Number of proposals (AN) as AR@AN, and calculate the Area under the AR vs. AN curve (AUC) as metrics, where AN is varied from 0 to 100.
	As can be seen in Table \ref{tab:activitynet_finetune}, our VCLR learned representations not only outperforms previous methods by a large margin on classification task, it also performs better on action localization. Qualitatively, we show two visualizations of our predicted proposals in Fig.~\ref{fig:localization}.

	\subsection{Downstream video retrieval}
	\label{subsec:downstream_video_retrieval}
	In addition to action classification and localization, we also evaluate VCLR on a common downstream task, video retrieval. For this task, we use the extracted feature from each video to perform nearest-neighbour retrieval, and the goal is to test if the query clip instance and its nearest neighbours belong to same semantic category.
	Similar to linear evaluation in Sec. \ref{subsec:k400}, we treat the pretrained model as a feature extractor, and no finetuning is performed.
	
\begin{figure*}[t]
		\centering
		\subfloat[]{
			\includegraphics[width=1  \columnwidth]{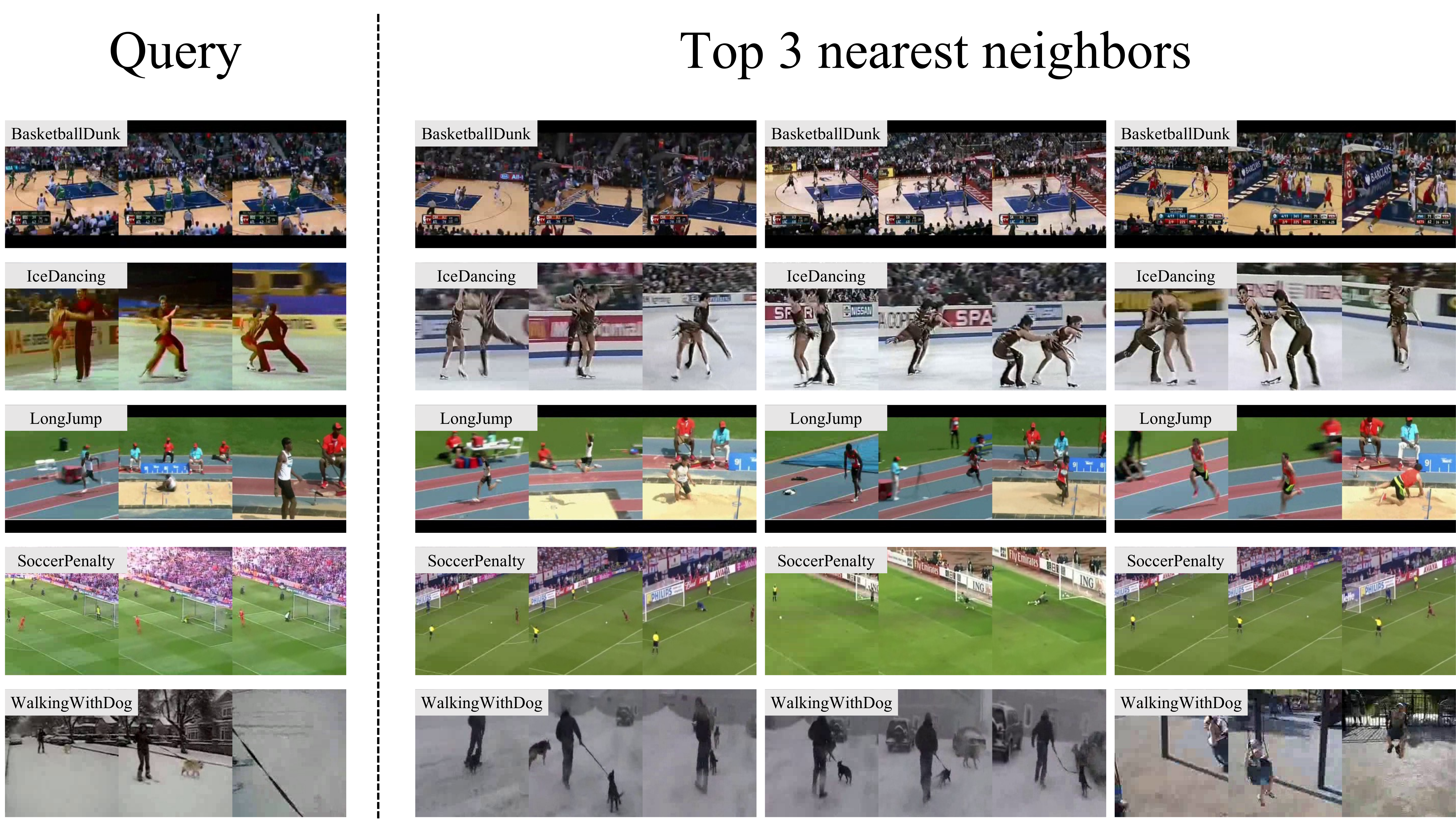}
			\label{fig:retrieval}
		}
		\hspace{2.2mm}
		\subfloat[]{
			\includegraphics[width=1  \columnwidth]{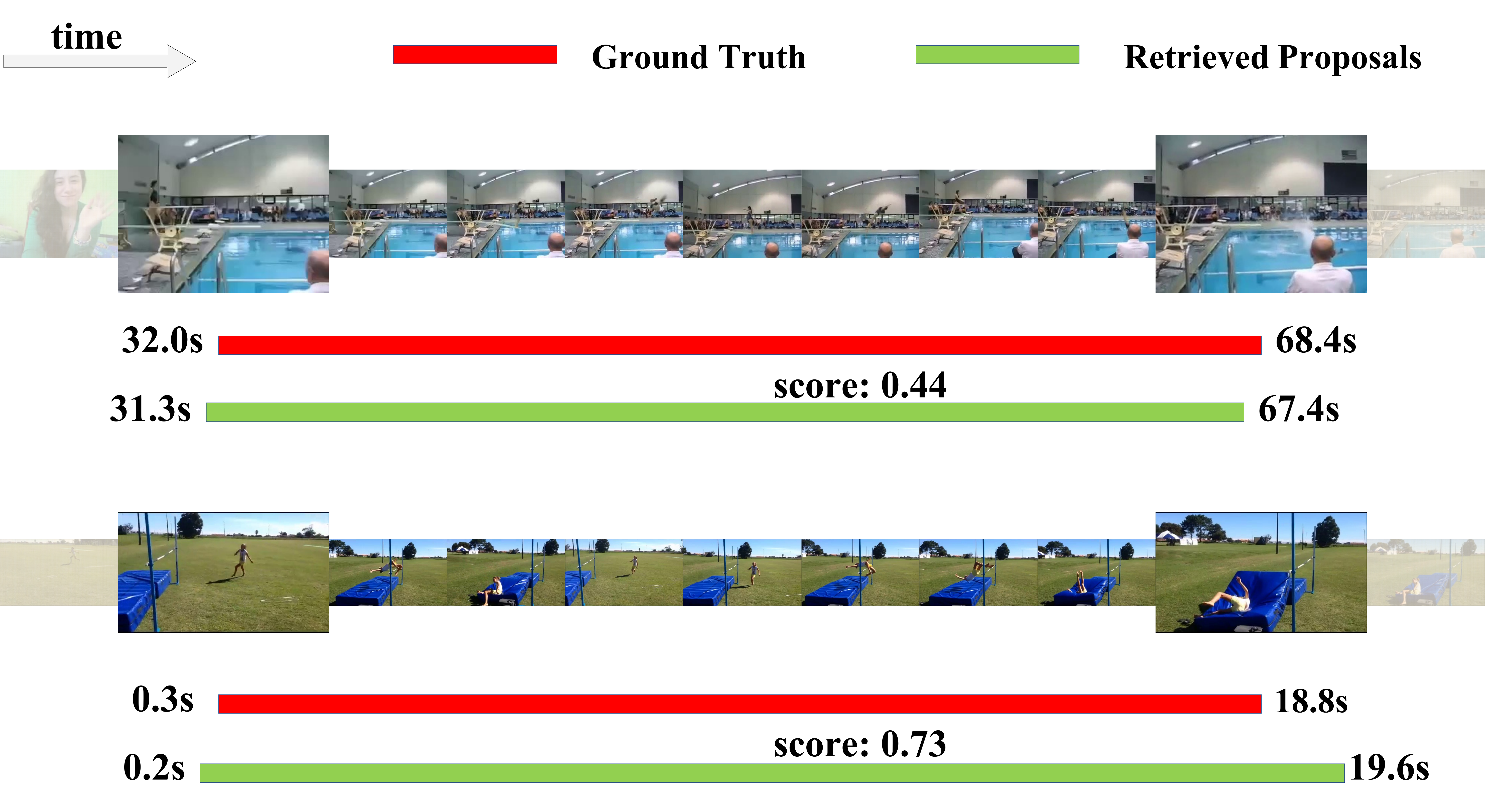}
			\label{fig:localization}
		}
		\vspace{-1ex}
		\caption{(a) \textbf{Video retrieval} results on UCF101. (b) \textbf{Action localization} results on ActivityNet. For both experiments, we use the learned representations from our VCLR model pretrained on Kinetics400 dataset.}
		\label{fig:goodresults}
		\vspace{-2ex}
	\end{figure*}

	Following \cite{xu_cvpr2019_vcop,han_nips2020_coclr}, we use the validation video samples in UCF101 and HMDB51 to search the $k$ nearest video samples from their training sets, respectively. We use Recall at $k$ (R$@k$) as evaluation metric, that means if one of the top $k$ nearest neighbours is from the same class as query, it is a correct retrieval. As can be seen in Table \ref{tab:ucf_hmdb_retrieval}, our VCLR trained representation outperforms other methods with respect to all the R$@k$ regardless of the pretraining dataset. 
	Qualitatively, we show several visualizations of our top-3 video retrievals on UCF101 dataset in Fig.~\ref{fig:retrieval}.
	We note that a recent work CoCLR \cite{han_nips2020_coclr} performs better than us. However, CoCLR has been trained with a heavier network S3D \cite{xie_eccv2018_s3d} and multiple hard positive samples mined by using optical flow. We conduct another experiment in Sec. \ref{subsec:generalization} to fairly compare with CoCLR using the same network architecture, and find that our performance is competitive to theirs even without using optical flow.

	\begin{table}[t]
		\centering
		\scalebox{0.8}{
			\begin{tabular}{cccc|c}
				\hline
				$\mathcal{L}_{\textit{Intra}}$ & $\mathcal{L}_{\textit{Inter}}$ & $\mathcal{L}_{\textit{Segment}}$ & $\mathcal{L}_{\textit{Order}}$ & Top-1 Acc. (\%) \\ 
				\hline
				\checkmark &  & & &  NA \\
				& \checkmark & & &  59.0 \\
				&  & \checkmark & &  62.1 \\
				&  & & \checkmark &  51.8 \\
				\hline
				\checkmark & \checkmark & & &  60.7 \\
				\checkmark & \checkmark &  & \checkmark &  61.2 \\
				\checkmark & \checkmark & \checkmark & &  63.5 \\
				\checkmark & \checkmark & \checkmark &  \checkmark &  \textbf{64.1} \\
				\hline
			\end{tabular}
		}
		\caption{\textbf{Ablation study on loss objectives}. We prertain all configurations on Kinetics400 under the same setting, and report linear evaluation performance on it. }
		\label{tab:loss_ablation}
		\vspace{-2ex}
	\end{table}
	
	\vspace{-2ex}
	\begin{table}[t]
		\centering
		\scalebox{0.8}{
			\begin{tabular}{c|c}
				\hline
				Num. of Segments ($K$)    & Top-1 Acc. (\%) \\ 
				\hline
				$K=1$        & 61.2             \\
				$K=2$        & 63.0             \\
				$K=3$        & 64.1             \\
				$K=4$        & 64.3             \\
				$K=5$        & 64.4             \\
				\hline
			\end{tabular}
		}
		\caption{\textbf{Ablation study on the number of segments}. We report the linear evaluation performance here.}
		\label{tab:num_seg}
		\vspace{-2ex}
	\end{table}
	
	\begin{table}[t]
		\centering
		\scalebox{0.8}{
			\begin{tabular}{l|c|c|cc}
				\hline
				\multirow{2}{*}{Method} &  
				\multirow{2}{*}{Modality} &
				\multirow{2}{*}{Retrieval} &
				\multicolumn{2}{c}{Classification} \\
				&  &  & Linear    & Finetune    \\
				\hline
				$\mathcal{L}_{\textit{Inter}}$ \cite{han_nips2020_coclr}  & V   &  33.1 & 46.8                  & 78.4 \\ 
				$\mathcal{L}_{\textit{Inter}}$ + $\mathcal{L}_{\textit{Segment}}$  & V   &  49.8 & 67.4                  & 80.2 \\ 
				$\mathcal{L}_{\textit{Inter}}$ + $\mathcal{L}_{\textit{Segment}}$ + $\mathcal{L}_{\textit{Order}}$  & V   &  \textbf{51.7} &      \textbf{70.9}             & \textbf{81.7} \\
				\hline
				\textcolor{gray}{CoCLR \cite{han_nips2020_coclr}} & \textcolor{gray}{VF}   &  \textcolor{gray}{51.8} &      \textcolor{gray}{70.2}             & \textcolor{gray}{81.4} \\
			\end{tabular}
		}
		\caption{\textbf{VCLR with 3D CNNs}. All models are pretrained on UCF101 under the same setting following \cite{han_nips2020_coclr}. }
		\label{tab:3dcnn}
		\vspace{-2ex}
	\end{table}

	\begin{table}[t]
		\centering
		\scalebox{0.8}{
			\begin{tabular}{l|c|cc}
				\hline
				\multirow{2}{*}{Method} &  
				\multirow{2}{*}{Pretrain} &
				\multicolumn{2}{c}{Classification} \\
				&   & UCF101    & HMDB51    \\
				\hline
				DPC \cite{han_iccvw2019_dpc} & UCF101 &      60.6           & 44.9 \\
				DPC + $\mathcal{L}_{\textit{Segment}}$     &  UCF101 &     62.0             & 46.3 \\
				DPC + $\mathcal{L}_{\textit{Segment}}$ + $\mathcal{L}_{\textit{Order}}$    &  UCF101 &     \textbf{ 62.7}             & \textbf{47.1} \\
			\end{tabular}
		}
		\caption{\textbf{VCLR with DPC} \cite{han_iccvw2019_dpc}. All models are pretrained under the same setting following \cite{han_iccvw2019_dpc}. }
		\label{tab:dpc}
	\end{table}

	\begin{table}[t]
	\centering
	\scalebox{0.8}{
	\begin{tabular}{l|c|cc}
	\hline
	\multirow{2}{*}{Method} &  
	\multirow{2}{*}{Pretrain} &
	\multicolumn{2}{c}{Classification} \\
	     &   & UCF101    & HMDB51    \\
	\hline
	SeCo \cite{yao_aaai2021_seco} & HACS &      64.1           & 47.2 \\
	PacePred \cite{wang_eccv2020_pace}    &  HACS &     62.8             & 46.6 \\
	DPC \cite{han_iccvw2019_dpc}    &  HACS &     60.1             & 45.4 \\
	\hline
	VCLR    &  HACS &     \textbf{67.2}             & \textbf{49.3} \\
	
	\end{tabular}
	}
	\caption{\textbf{Pretrain on multi-label dataset HACS} \cite{zhao_arxiv2019_hacs}. All methods use R3D-18 for fair comparison.}
	\label{tab:multilabel}
	\vspace{-2ex}
	\end{table}

	\section{Discussion}
	\label{sec:discussion}
	In this section, we present ablation studies and important discussions on VCLR. Given space limitation, we put more experiments, visualizations and implementation details in the supplemental materials.

	\subsection{Ablation study on loss objectives}
	\label{subsec:ablation_loss}
	Our method VCLR has four loss objectives:  $\mathcal{L}_{\textit{Intra}}$, $\mathcal{L}_{\textit{Inter}}$, $\mathcal{L}_{\textit{Segment}}$ and  $\mathcal{L}_{\textit{Order}}$. We now dissect their contributions and see how they impact the learned representations.
	
	We have several observations from Table \ref{tab:loss_ablation}. First, video-level instance discrimination $\mathcal{L}_{\textit{Segment}}$ is a strong supervision signal. By using this loss alone could learn good representations and outperform previous state-of-the-art SeCo ($62.1$ vs $61.9$). Second, temporal order constraint $\mathcal{L}_{\textit{Order}}$ is relatively weak. Using it alone only achieves $51.8\%$ accuracy. However, when combined with other loss objectives, it can provide further regularization and push the performance to $64.1$. Third, using intra-frame instance discrimination $\mathcal{L}_{\textit{Intra}}$ alone is not enough to stabilize training as the negative set is too small to perform contrasting.

	\subsection{Ablation study on the number of segments}
	\label{subsec:ablation_segments}
	We set $K=3$ as default number of segments in our experiments, now we discuss the impact of the number of segments for our method. We present the comparisons with respect to linear evaluation performance on Kinetics400 dataset in Table~\ref{tab:num_seg}. 
	
	We can see that with the increasing number of segments, the performance continue to improve, which indicates the necessity of using global context in training video models. However, the performance starts to saturate when using more segments.
	In our paper, we simply choose $K$=$3$ for a better training speed-accuracy trade-off.

	\subsection{General applicability}
	\label{subsec:generalization}
	Our framework VCLR is not limited to any network architecture. For example, we could sample short video clips instead of frames from each segment to formulate positive pairs. In this way, a 3D CNN can be pretrained without using labels. In addition, VCLR is compatible with most loss objectives, and is orthogonal to other advancements in self-supervised video representation learning.
	
	\noindent \textbf{VCLR with 3D CNNs}
	Here we train a R3D-50 model using VCLR to demonstrate our framework's general applicability. In terms of baseline, we compare to a R3D-50 model pretrained by $\mathcal{L}_{\textit{Inter}}$ on UCF101 dataset following \cite{han_nips2020_coclr}. We then use three segments and pretrain the model by adding loss objectives $\mathcal{L}_{\textit{Segment}}$ and $\mathcal{L}_{\textit{Order}}$. As seen in Table \ref{tab:3dcnn}, our method significantly outperforms the corresponding baseline with respect to both video classification and retrieval task on UCF101. In addition, we compare to CoCLR \cite{han_nips2020_coclr}, which uses optical flow to mine multiple hard positives for contrastive learning. We can see that our method is competitive to CoCLR even without optical flow.
	
	\noindent \textbf{VCLR with DPC}
	Both $\mathcal{L}_{\textit{Segment}}$ and $\mathcal{L}_{\textit{Order}}$ are compatible to other algorithms in self-supervised video representation learning. Similarly, we incorporate them to a recent work DPC \cite{han_iccvw2019_dpc}, which uses dense prediction as a pretext task, to train a R3D-18 model on UCF101. 
	As shown in Table \ref{tab:dpc}, a simple addition improves the transfer performance to downstream action classification task on both UCF101 and HMDB51, which demonstrates the necessity of using global context. 
	We believe VCLR can be combined with other methods \cite{yang_arxiv2020_vthcl,qian_cvpr2021_cvrl} and show performance improvements.

	\subsection{Visualization}
	\label{subsec:visualization}
\begin{figure}[h!]
		\begin{center}
			\includegraphics[width=1.0 \columnwidth]{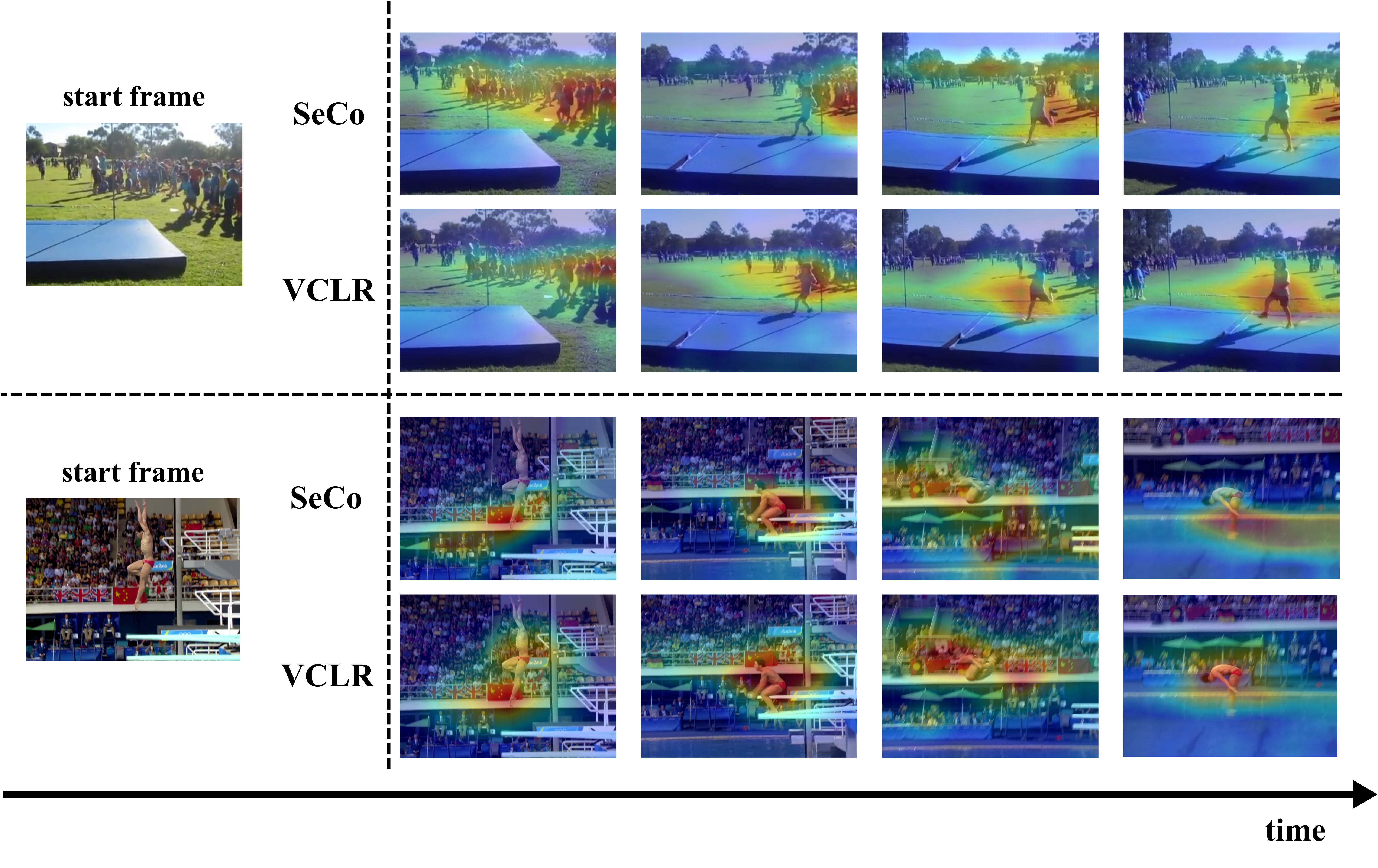}
		\end{center}
		\vspace{-4ex}
		\caption{\textbf{Attention visualization between SeCo \cite{yao_aaai2021_seco} and VCLR.} The upper video is from Kinetics400 validation set, and the bottom one is an unseen video from Internet.}
		\label{fig:feat_map}
		\vspace{-2ex}
	\end{figure}

	We visualize the activations of the conv5 output from ResNet50 models pretrained by SeCo \cite{yao_aaai2021_seco} and ours. As can be seen in Fig. \ref{fig:feat_map}, VCLR is able to focus on the moving region of interest, while SeCo focuses more on the background (e.g., springboard and water). It is also interesting to note that VCLR generalizes better to unseen videos.
	
	\subsection{Pretrain on multi-label videos}
	\label{subsec:multilabel}
	As aforementioned, our framework is able to generate robust positive pairs for contrastive learning compared with clip-level sampling strategy. However, we conduct all experiments by pretraining the model on Kinetics400 dataset in order to perform fair comparison with previous literature. Here we pretrain the model on a multi-label dataset, HACS \cite{zhao_arxiv2019_hacs}, to demonstrate the advantage of VCLR. 
	
	We compare VCLR to three methods, SeCo \cite{yao_aaai2021_seco}, PacePred \cite{wang_eccv2020_pace} and DPC \cite{han_iccvw2019_dpc}. They correspond to the three different ways of formulating positive pairs as mentioned in Sec. \ref{sec:introduction}, namely instance-based, pace-based and prediction-based contrastive learning. We use the same backbone R3D-18, and perform finetuning on UCF101 to evaluate the quality of learned representations. As seen in Table \ref{tab:multilabel}, our VCLR outperforms them by a large margin, which indicates the importance of generating correct positive pairs and the necessity of using global context.

	\section{Conclusion}
	\label{sec:conclusion}
	In this paper, we introduce a new way to formulate positive pairs for video-level contrastive learning. We also incorporate a temporal order regularization to enforce the inherent sequential structure of videos during training. Our final framework VCLR shows state-of-the-art results on five video datasets for downstream action recognition/localization and video retrieval. Furthermore, we demonstrate the generalization capability of our method, e.g., pretrain 3D CNNs and compatible with other self-supervised learning algorithms.
	We want to emphasize that the effectiveness of VCLR is beyond the experimental results presented in this paper due to using trimmed video datasets for the sake of fair comparison. Our method is applicable to various inputs (short trimmed or long untrimmed videos), networks (2D or 3D CNNs), methods (various algorithms) and ready to scale to future datasets for video self-supervised representation learning.
	We hope VCLR can serve as a strong baseline and facilitate research of using global context for video understanding.

	
	{\small
		\bibliographystyle{ieee_fullname}
		\bibliography{egbib}
	}
	

	\newpage
	
	\appendix
	
	
	\normalsize In the appendix, we provide more implementation details, performance comparisons and visualizations. Specifically, we first describe the implementation details of experiments on pretraining, downstream tasks and discussions in Sec.~\ref{sec:supp_implement}. We then present a complete comparison to other published methods in terms of downstream action classification in Sec.~\ref{sec:supp_downstream}. In the end, we show more visualizations of our successful predictions, failure cases and feature visualizations in Sec.~\ref{sec:supp_vis}.
	
	\section{Implementation Details}
	\label{sec:supp_implement}
	
	\subsection{Pretraining setup}
	Our query encoder $f_{q}$ is a ResNet50 model. Following \cite{chen_arxiv2020_mocov2,yao_aaai2021_seco}, we also have a key encoder $f_{k}$ (a.k.a, momentum encoder) which has the same architecture as $f_{q}$ but updated with a momentum strategy,
	\begin{equation}
	f_{k}^{t} = m f_{k}^{t-1} + (1 - m) f_{q}^{t-1}.
	\label{eq:momentum_update}
	\end{equation}
	Here, $t$ indicates the training iteration step. We set the momentum $m$ to $0.999$. 
	The anchor samples are always encoded by query encoder $f_{q}$. Both positive and negative samples are encoded by key encoder $f_{k}$. 
	The negative samples are stored in a memory bank. We have two memory banks, one for video-level contrastive loss $\mathcal{L}_{\textit{Segment}}$, the other for inter-frame contrastive loss $\mathcal{L}_{\textit{Inter}}$. Both memory banks have a size of $131,072$, following SeCo \cite{yao_aaai2021_seco}. 
	Since we have four loss objectives, we have four MLP heads after the backbone encoder. All four MLP heads have the same architecture, but do not share weights. Specifically, each head consists of a fully-connected layer ($ 2048 \times 2048 $), a ReLU activation layer and a final embedding layer ($ 2048 \times 128 $).
	All the embeddings are then L2-normalized before computing the loss objectives. The overall loss is a simple summation over four losses. We did not do parameter tuning on the loss weights, although weighted summation might lead to improved performance.
	We note that these four MLP heads are only used during training, but discarded for downstream tasks.
	In terms of optimization, we train the model on Kinetics400 dataset with a batch size of 512 for 400 epochs. We also employ the Shuffle-BN for multi-GPU training as in \cite{he_cvpr2020_moco}. We use SGD to optimize the network with an initial learning rate of $0.2$, and annealed to zero with a cosine decay scheduler.
	During linear evaluation, the ResNet50 backbone is frozen and treated as a feature extractor. We train a linear SVM~\footnote{\url{https://www.csie.ntu.edu.tw/~cjlin/liblinear/}} with the Kinetics400 training set and report performance on its validation set.

	\subsection{Downstream setup}
	
	For all the downstream tasks, we adopt mmaction2 \cite{2020mmaction2} toolkit and mostly follow their default settings to perform finetuning on each dataset. 
	
	\subsubsection{Action classification}
	
	\paragraph{UCF101}
	We use the pretrained ResNet50 as backbone network initialization and finetune it on UCF101 dataset. During training, we uniformly divide each video into 3 segments, and select one frame from each segment. For data augmentation, we apply random resized cropping, random horizontal flip, and then resize it to $ 224 \times 224 $. We use average operation for taking consensus $\mathcal{C}$ following TSN~\cite{wang_eccv2016_tsn}. We use SGD to optimize the model for $300$ epochs with a batch size of $256$. The initial learning rate is set to $ 0.01 $, and decayed by $ 0.1 $ at $120$ and $240$ epoch respectively. The momentum is $ 0.9 $ and the weight decay is $ 5 \times 10^{-4} $. During test, we uniformly sample 25 frames from each video, perform center crop to $224 \times 224$ and average their predictions to be the final video prediction.
	
	\paragraph{HMDB51}
	We use the pretrained ResNet50 as backbone network initialization and finetune it on HMDB51 dataset. During training, we uniformly divide each video into 8 segments, and select one frame from each segment. For data augmentation, we apply random resized cropping, random horizontal flip, and then resize it to $ 224 \times 224 $. We use average operation for taking consensus $\mathcal{C}$ following TSN~\cite{wang_eccv2016_tsn}. We use SGD to optimize the model for $50$ epochs with a batch size of $256$. The initial learning rate is set to $ 0.025 $, and decayed by $ 0.1 $ at $20$ and $40$ epoch respectively. The momentum is $ 0.9 $ and the weight decay is $ 1 \times 10^{-4} $. During test, we uniformly sample 8 frames from each video, perform center crop to $224 \times 224$ and average their predictions to be the final video prediction.
	
	\paragraph{SthSthV2} 
	We use the pretrained ResNet50 as backbone network initialization and finetune it on SthSthV2 dataset. During training, we uniformly divide each video into 8 segments, and select one frame from each segment. For data augmentation, we apply multi-scale cropping with scales $ \{1, 0.875, 0.75, 0.066\} $, random horizontal flip, and then resize it to $ 224 \times 224 $. We use average operation for taking consensus $\mathcal{C}$ following TSN~\cite{wang_eccv2016_tsn}. We use SGD to optimize the model for $50$ epochs with a batch size of $128$. The initial learning rate is set to $ 0.02 $, and decayed by $ 0.1 $ at $20$ and $40$ epoch respectively. The momentum is $ 0.9 $ and the weight decay is $ 1 \times 10^{-4} $. During test, we uniformly sample 8 frames from each video, perform 10-crop to $224 \times 224$ and average their predictions to be the final video prediction.
	
	We also finetune a recent  TSM~\cite{lin_iccv2019_tsm} model for action classification on SthSthV2. We only change the initial learning rate to $ 0.0075 $ , weight decay to $ 5 \times 10^{-4}$, batch size to $ 48 $, the rest setting remain the same as TSN training.
	
	\paragraph{ActivityNet} 
	We use the pretrained ResNet50 as backbone network initialization and finetune it on ActivityNet dataset. During training, we uniformly divide each video into 8 segments, and select one frame from each segment. For data augmentation, we apply random resized cropping, random horizontal flip, and then resize it to $ 224 \times 224 $. We use average operation for taking consensus $\mathcal{C}$ following TSN~\cite{wang_eccv2016_tsn}. We use SGD to optimize the model for $50$ epochs with a batch size of $64$. The initial learning rate is set to $ 0.01 $, and decayed by $ 0.1 $ at $20$ and $40$ epoch respectively. The momentum is $ 0.9 $ and the weight decay is $ 1 \times 10^{-4} $. During test, we uniformly sample 25 frames from each video, perform 3-crop to $224 \times 224$ and average their predictions to be the final video prediction.
	
	\subsubsection{Action localization}
	For action localization on ActivityNet, we first extract video features from the finetuned ResNet50 model on ActivityNet mentioned above. We then use them to train a recent BMN~\cite{lin_iccv2019_bmn} model to perform action localization. The BMN model is optimized by Adam for 9 epochs with a batch size of 16. The initial learning rate is set to $ 0.001 $ and decayed by $ 0.1 $ at epoch 7. 
	
	\subsubsection{Video retrieval}
	For video retrieval, we use the validation video samples in UCF101 and HMDB51 to search the top k nearest neighbors from training set following ~\cite{xu_cvpr2019_vcop, luo_aaai2020_vcp}. In details, we sample 30 frames of each video, then input them to ResNet50 and employ consensus $\mathcal{C}$ with average operation to get the video features. Then, we search the k nearest neighbors according to the cosine similarity between input video features with each training video.

	\subsection{Discussion setup}
	
	\paragraph{VCLR with 3D CNNs}
	Following CoCLR~\cite{han_nips2020_coclr}, we choose the S3D architecture~\cite{xie_eccv2018_s3d} as the backbone for all three experiments. For $\mathcal{L}_{\textit{Inter}}$, we use 32-frame video clip as input. When adding the other two losses, we uniformly divide the video into 3 segments and randomly pick a 32-frame video clip from each segment. Each frame is of size $128 \times 128$. We train the model on UCF101 for 300 epochs with a batch size of 64. The initial learning rate is set to $0.001$ and decayed by 10 at epoch 250 and 280.
	
	\paragraph{VCLR with DPC}
	Following DPC \cite{han_iccvw2019_dpc}, we choose R3D-18 architecture as the backbone for all three experiments.
	When adding the other two losses, we uniformly divide the video into 3 segments and randomly pick a 5-frame video clip from each segment. Each frame is of size $128 \times 128$. We train the model on UCF101 for 300 epochs with a batch size of 128. The initial learning rate is set to $0.001$ and decayed by 10 at epoch 150 and 250.
	
	\begin{table*}[ht]
		\centering
		\scalebox{1}{
			\begin{tabular}{lr|c|c|c|c|cc}
				\hline
				\multirow{2}{*}{Method} & \multirow{2}{*}{Venue} & \multirow{2}{*}{\begin{tabular}[c]{@{}c@{}}Pretraining\\dataset (duration) \end{tabular}} & \multirow{2}{*}{Network} & \multirow{2}{*}{Mod.} & \multirow{2}{*}{Frozen} & \multicolumn{2}{c}{Top-1 Acc (\%)} \\
				&    								&														   &  									  & 				&							& UCF101  		& HMDB51        \\ 
				\hline
				MemDPC \cite{han_eccv2020_memdpc}				& ECCV2020						  	& K400 (28d)					  	& R-2D3D 				 & V		  & \checkmark 	 & 54.1         		 & 30.5        \\
				\textcolor{gray}{XDC \cite{alwassel_2020_xdc}}		& \textcolor{gray}{NeurlPS 2020}						  	& \textcolor{gray}{IG65M (21y)}					  	& \textcolor{gray}{R(2+1)D} 				 & \textcolor{gray}{VA}		  & \textcolor{gray}{\checkmark} 	 & \textcolor{gray}{85.3}      		 & \textcolor{gray}{56.1}        \\
				\textcolor{gray}{MIL-NCE \cite{miech_2020_milnce}}			& \textcolor{gray}{CVPR 2020}						  	& \textcolor{gray}{HTM (15y)}					  	& \textcolor{gray}{S3D} 				 & \textcolor{gray}{VT}		  & \textcolor{gray}{\checkmark} 	 & \textcolor{gray}{82.7}      		 & \textcolor{gray}{53.1}        \\
				\textcolor{gray}{MIL-NCE \cite{miech_2020_milnce}}			& \textcolor{gray}{CVPR 2020}						  	& \textcolor{gray}{HTM (15y)}					  	& \textcolor{gray}{I3D} 				 & \textcolor{gray}{VT}		  & \textcolor{gray}{\checkmark} 	 & \textcolor{gray}{83.4}      		 & \textcolor{gray}{54.1}        \\
				\textcolor{gray}{Elo	\cite{piergiovanni_2020_elo}}		& \textcolor{gray}{CVPR 2020}						  	& \textcolor{gray}{Youtube8M (8y)}					  	& \textcolor{gray}{I3D} 				 & \textcolor{gray}{VAF}		  & \textcolor{gray}{\checkmark} 	 & \textcolor{gray}{-}     		 & \textcolor{gray}{64.5}        \\
				\textcolor{gray}{CoCLR \cite{han_nips2020_coclr}}			& \textcolor{gray}{NeurlPS 2020}						  	& \textcolor{gray}{K400 (28d)}					  	& \textcolor{gray}{S3D} 				 & \textcolor{gray}{VF}		  & \textcolor{gray}{\checkmark} 	 & \textcolor{gray}{77.8}     		 & \textcolor{gray}{52.4}        \\
				\hline
				VCLR                    			& 							  			& K400 (28d)	         			 	& R2D-50 			& V				& \checkmark 		& \textbf{85.1}        & \textbf{51.6}       \\ 
				\hline \hline
				MAS \cite{wang_cvpr2019_statistics}					& CVPR 2019						  	& K400 (28d)					  	& C3D 				 & V		  & \xmark 	 & 61.2         		 & 33.4        \\
				ST-Puzzle \cite{kim_aaai2019_cubicPuzzle}		& AAAI 2019				  			& K400 (28d)						& R3D-50			& V 		& \xmark	& 65.8                  & 33.7 \\ 
				VCOP \cite{xu_cvpr2019_vcop}						 & CVPR 2019       					 & K400 (28d)	   				 	& R(2+1)D			 & V   		& \xmark	 & 72.4                  & 30.9        \\
				DPC \cite{han_iccvw2019_dpc}						 & ICCVW 2019                   		 & K400 (28d)	  				 & R-2D3D 			& V			 	 & \xmark	 & 75.7                  & 35.7        \\
				DynamoNet \cite{diba_iccv2019_dynamoNet}						 & ICCV 2019                   		 & Youtube8M-1 (58d)	  				 & STCNet 			& V			 	 & \xmark	 & 88.1                 & 59.9       \\
				PacePred \cite{wang_eccv2020_pace}				  & ECCV 2020              			  & K400 (28d)	  				  & R(2+1)D  		 & V			  & \xmark		& 77.1                  & 36.6        \\
				MemDPC \cite{han_eccv2020_memdpc}			  & ECCV 2020      					  & K400 (28d)				   	  & R-2D3D  		& V 			 & \xmark	 & 78.1                  & 41.2        \\
				TT \cite{jenni_eccv2020_temporalssl}				& ECCV 2020              			& K400 (28d)	 					& R3D-18  		  & V			& \xmark	 & 79.3                  & 49.8        \\
				VIE \cite{zhuang_cvpr2020_VIE}						  & CVPR 2020    					  & K400 (28d)		   			  & SlowFast 		 & V			  & \xmark 		& 80.4                  & 52.5        \\
				SpeedNet \cite{benaim_cvpr2020_speednet}	  & CVPR 2020    			 		  & K400	(28d)				      & S3D-G			  & V   		& \xmark	& 81.1                  & 48.8        \\
				Bi-Pred \cite{behrmann_wacv2021_bipred}	  			  & WACV 2021    					 & K400 (28d)		 			 & R2D-50 			& V				 & \xmark		& 66.4                  & 45.3       \\
				SeCo \cite{yao_aaai2021_seco}	  & AAAI 2021    					 & K400 (28d)		 			 & R2D-50 			& V				 & \xmark		& 83.4                  & 49.7       \\
				CVRL \cite{qian_cvpr2021_cvrl}	  & CVPR 2021    					 & K400 (28d)		 			 & R3D-50 			& V				 & \xmark		& \textbf{92.9}                  & \textbf{67.9}     \\
				\hline 
				\textcolor{gray}{MemDPC \cite{han_eccv2020_memdpc}} 		& ECCV 2020 		 & \textcolor{gray}{K400 (28d)} 		& \textcolor{gray}{R-2D3D} 		& \textcolor{gray}{VF} 		& \xmark 	& \textcolor{gray}{86.1} 		& \textcolor{gray}{54.5} \\
				\textcolor{gray}{CoCLR \cite{han_nips2020_coclr}} 		& NeurIPS 2020 		 & \textcolor{gray}{K400 (28d)} 		& \textcolor{gray}{S3D} 		& \textcolor{gray}{VF} 		& \xmark 		& \textcolor{gray}{90.6} 		& \textcolor{gray}{62.9} \\
				\textcolor{gray}{AVTS \cite{korbar_2018_avts}} 		& NeurIPS 2018 		 & \textcolor{gray}{AudioSet (240d)} 		& \textcolor{gray}{MC3} 		& \textcolor{gray}{VA} 		& \xmark 		& \textcolor{gray}{89.0} 		& \textcolor{gray}{61.6} \\
				\textcolor{gray}{XDC \cite{alwassel_2020_xdc}} 		& NeurIPS 2020 		 & \textcolor{gray}{AudioSet (240d)} 		& \textcolor{gray}{R(2+1)D} 		& \textcolor{gray}{VA} 		& \xmark 		& \textcolor{gray}{91.2} 		& \textcolor{gray}{61.0} \\	
				\textcolor{gray}{XDC \cite{alwassel_2020_xdc}} 		& NeurIPS 2020 		 & \textcolor{gray}{IG65M (21y)} 		& \textcolor{gray}{R(2+1)D} 		& \textcolor{gray}{VA} 		& \xmark 		& \textcolor{gray}{94.2} 		& \textcolor{gray}{67.4} \\	
				\textcolor{gray}{MIL-NCE \cite{miech_2020_milnce}}		& CVPR 2020 		 & \textcolor{gray}{HTM (15y)} 		& \textcolor{gray}{S3D} 		& \textcolor{gray}{VA} 		& \xmark 		& \textcolor{gray}{91.3} 		& \textcolor{gray}{61.0} \\
				\textcolor{gray}{Elo \cite{piergiovanni_2020_elo}} 		& CVPR 2020 		 & \textcolor{gray}{Youtube8M-2 (13y)} 		& \textcolor{gray}{R(2+1)D} 		& \textcolor{gray}{VAF} 		& \xmark 		& \textcolor{gray}{93.8} 		& \textcolor{gray}{67.4} \\
				\textcolor{gray}{MMV \cite{alayrac_nips2020_mmv}} 		& NeurlPS 2020 		 & \textcolor{gray}{AudioSet+HTM (16y)} 		& \textcolor{gray}{S3D} 		& \textcolor{gray}{VA} 		& \xmark 		& \textcolor{gray}{91.1} 		& \textcolor{gray}{68.3} \\	
				\textcolor{gray}{MMV \cite{alayrac_nips2020_mmv}} 		& NeurlPS 2020		 & \textcolor{gray}{AudioSet+HTM (16y)} 		& \textcolor{gray}{S3D} 		& \textcolor{gray}{VAT} 		& \xmark 		& \textcolor{gray}{95.2} 		& \textcolor{gray}{75.0} \\			
				\hline
				VCLR                    			& 							  			& K400 (28d)	         			 	& R2D-50 			& V				& \xmark 		& 85.6        & 54.1       \\ 
				\hline \hline
				\textcolor{gray}{Sup-ImageNet} 		&               			& -   			 		& \textcolor{gray}{R2D-50}				& \textcolor{gray}{V}        & \xmark 		& \textcolor{gray}{81.6}    & \textcolor{gray}{49.0}       \\ 
				\textcolor{gray}{Sup-K400} 		&               				  & -   				   & \textcolor{gray}{R2D-50}             & \textcolor{gray}{V} 	   & \xmark 		& \textcolor{gray}{88.1}     & \textcolor{gray}{56.1}       \\
				\textcolor{gray}{Sup-K400} 		&               				  & -   				   & \textcolor{gray}{R3D-50}             & \textcolor{gray}{V} 	   & \xmark 		& \textcolor{gray}{89.3}     & \textcolor{gray}{61.0}       \\
				\textcolor{gray}{Sup-K400} 		&               				  & -   				   & \textcolor{gray}{R(2+1)D}             & \textcolor{gray}{V} 	   & \xmark 		& \textcolor{gray}{96.8}     & \textcolor{gray}{74.5}       \\
				\textcolor{gray}{Sup-K400} 		&               				  & -   				   & \textcolor{gray}{S3D}             & \textcolor{gray}{V} 	   & \xmark 		& \textcolor{gray}{96.8}     & \textcolor{gray}{75.9}       \\
				\hline
			\end{tabular}
		}
		\caption{\textbf{Downstream action classification on UCF101 and HMDB51}.  Sup-Kinetics400: supervised Kinetics400 pretrained weights. V: RGB frames. F: optical flow. A: audio. T: text.}
		\label{tab:ucf_hmdb51_finetune_full}
		\vspace{-2ex}
	\end{table*}

	\section{Comprehensive comparison among methods in downstream action classification}
	\label{sec:supp_downstream}
	In the main submission, we present a smaller table for fair comparison to other methods on UCF101 and HMDB51. Here, we aim to cover recent literature as comprehensive as possible, such as using larger pretraining dataset (i.e., Youtube 8M, IG65M) and other modalities (i.e., optical flow, audio and text). 
	As shown in Table.~\ref{tab:ucf_hmdb51_finetune_full}, we can see that our proposed VCLR outperforms other methods in both settings (frozen and finetuning) and on both datasets (UCF101 and HMDB51). We have several observations. First, larger pretraining dataset and using multi-modalities usually gives better performance. Second, 3D CNNs achieve higher accuracy than 2D CNNs, regardless of in supervised or unsupervised learning regime. Hence, our future work would be incorporating VCLR into 3D CNNs or recent Transformer architectures for improved performance.

	\section{More visualizations}
	\label{sec:supp_vis}
	
	\subsection{Video retrieval visualization}
\begin{figure*}[t]
		\centering
		\subfloat[]{
			\includegraphics[height=1  \columnwidth]{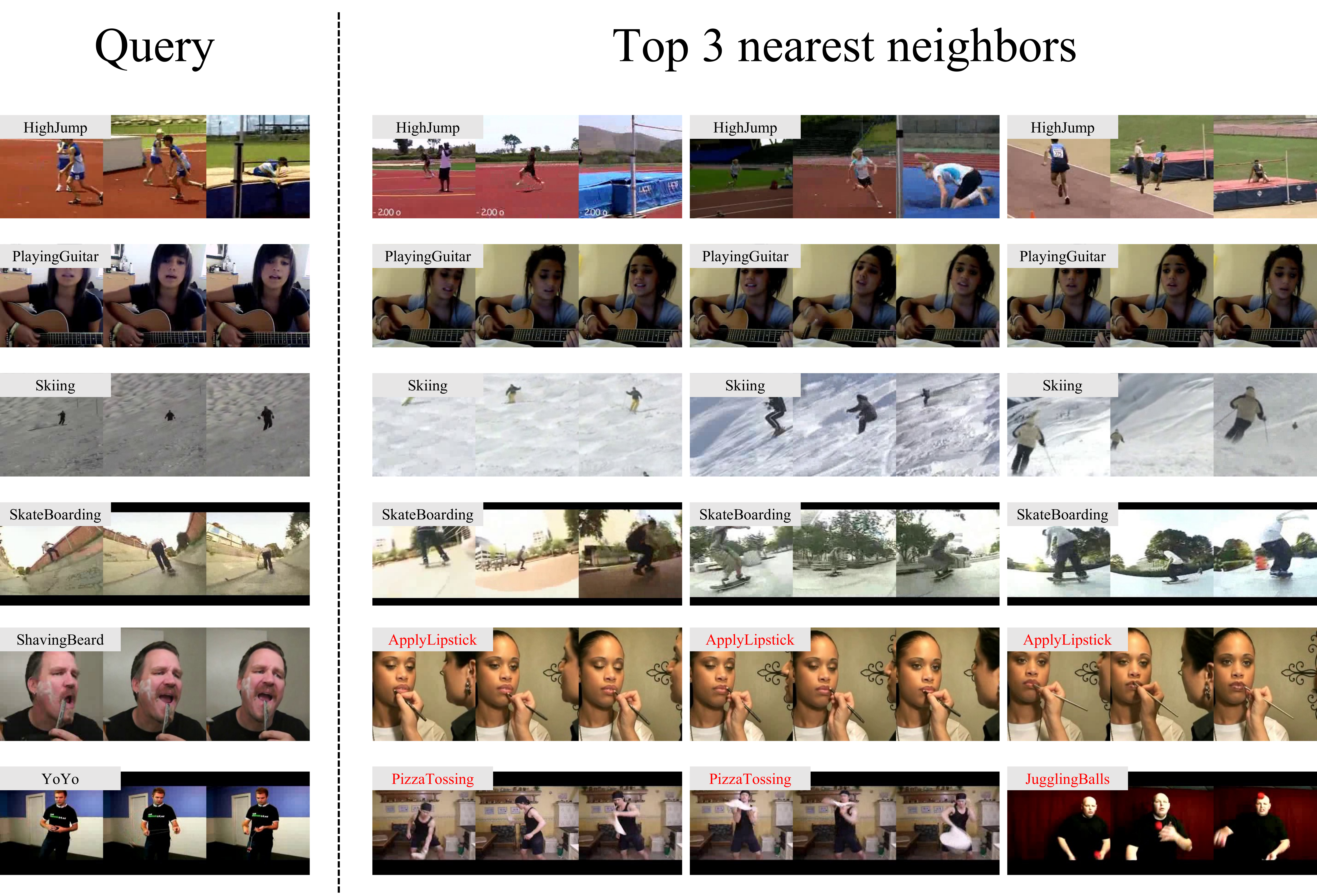}
			\label{fig:retrieval_ucf}
		}
		\quad
		\subfloat[]{
			\includegraphics[height=1  \columnwidth]{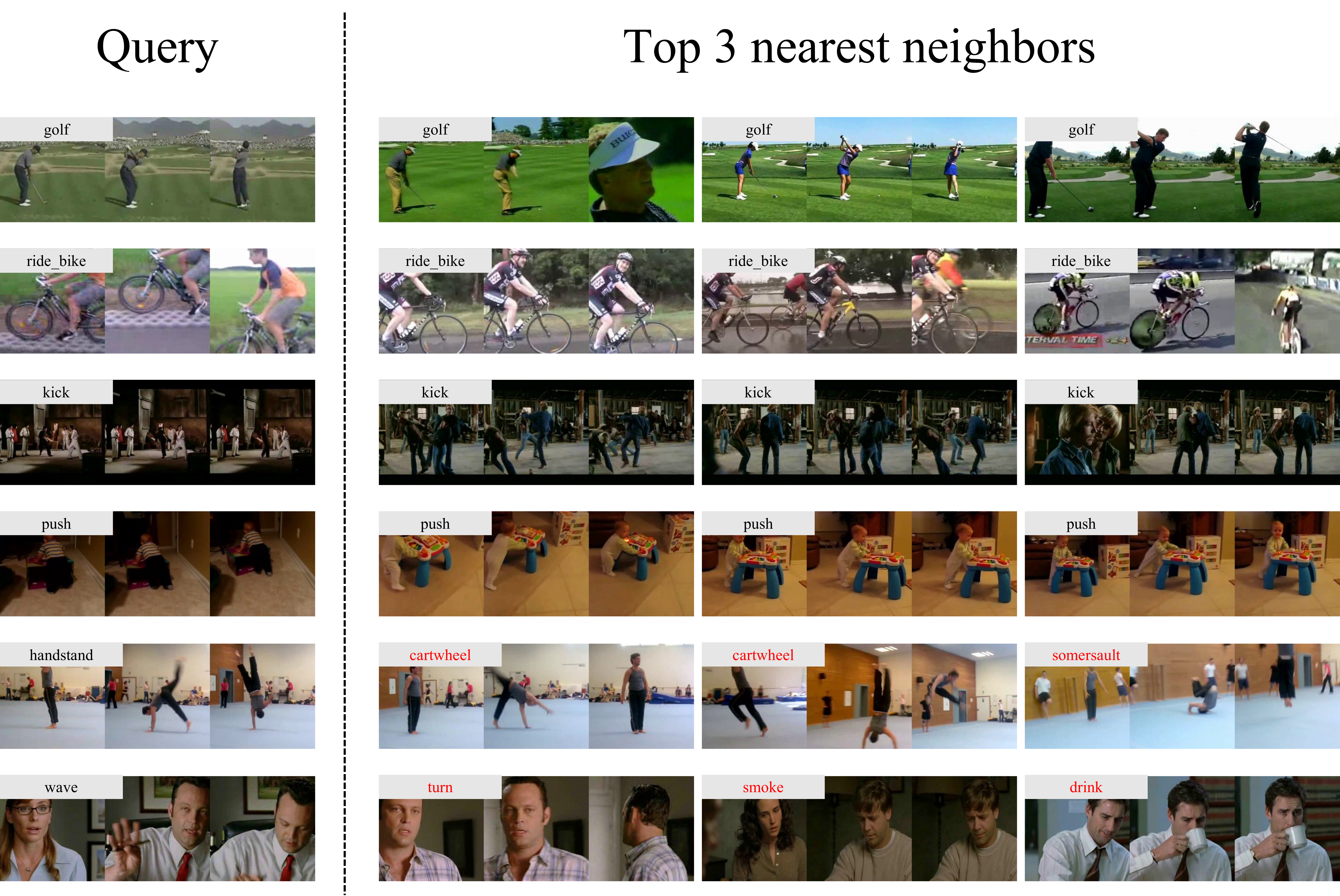}
			\label{fig:retrieval_hmdb}
		}
		\caption{\textbf{Downstream video retrieval results}. (a) Six examples of UCF101 dataset. (b) Six examples of HMDB51 dataset. Within each six examples, top four rows are correct retrievals and the last two rows are incorrect. }
		\label{fig:retrieval_supp}
	\end{figure*}

	In Fig.~\ref{fig:retrieval_supp}, we choose $k = 3$ to report top-3 retrieval results. We show both successful retrievals and failure cases. We can see that video frames retrieved by our VCLR method are highly correlated with the query. Even in the failure cases, both the appearance and motion patterns of different actions are very similar. 
	
	\subsection{Action localization visualization}
\begin{figure*}[t]
		\centering
		\subfloat[]{
			\includegraphics[width=1  \columnwidth]{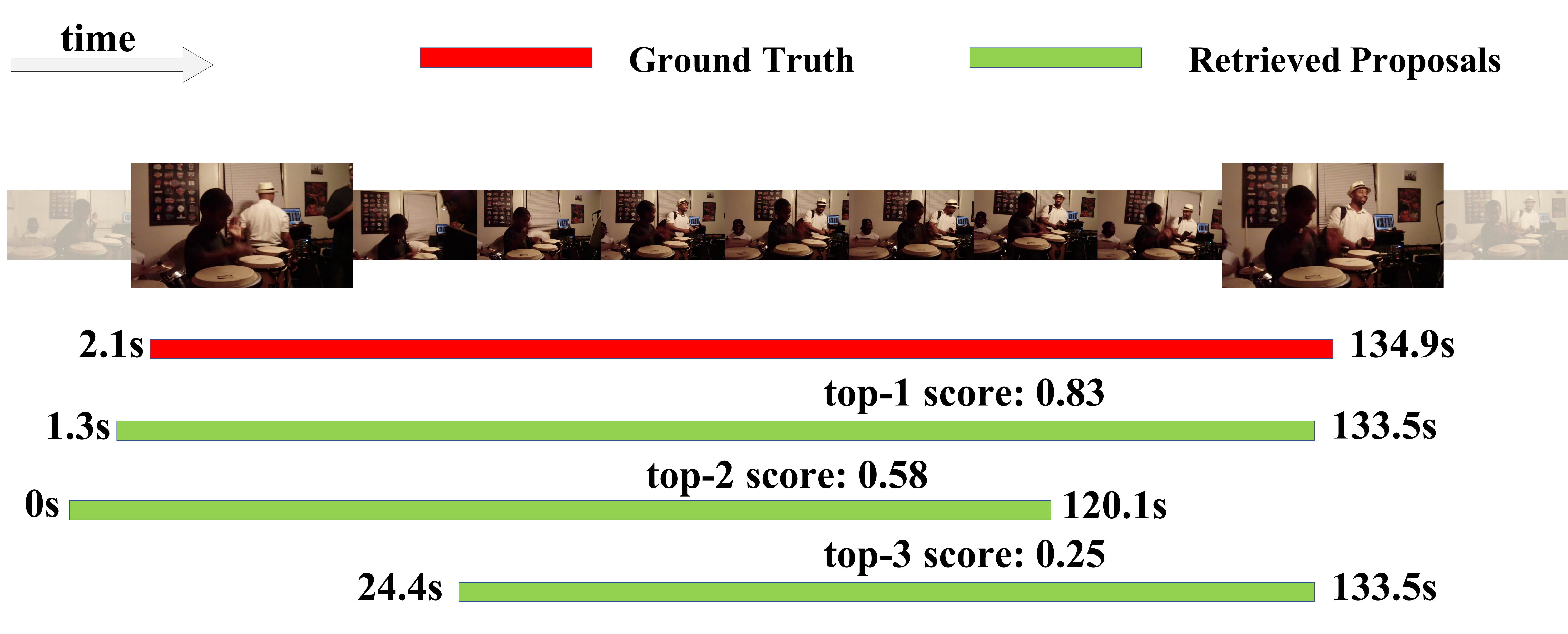}
			\label{fig:localization_a}
		}
		\hspace{2.2mm}
		\subfloat[]{
			\includegraphics[width=1  \columnwidth]{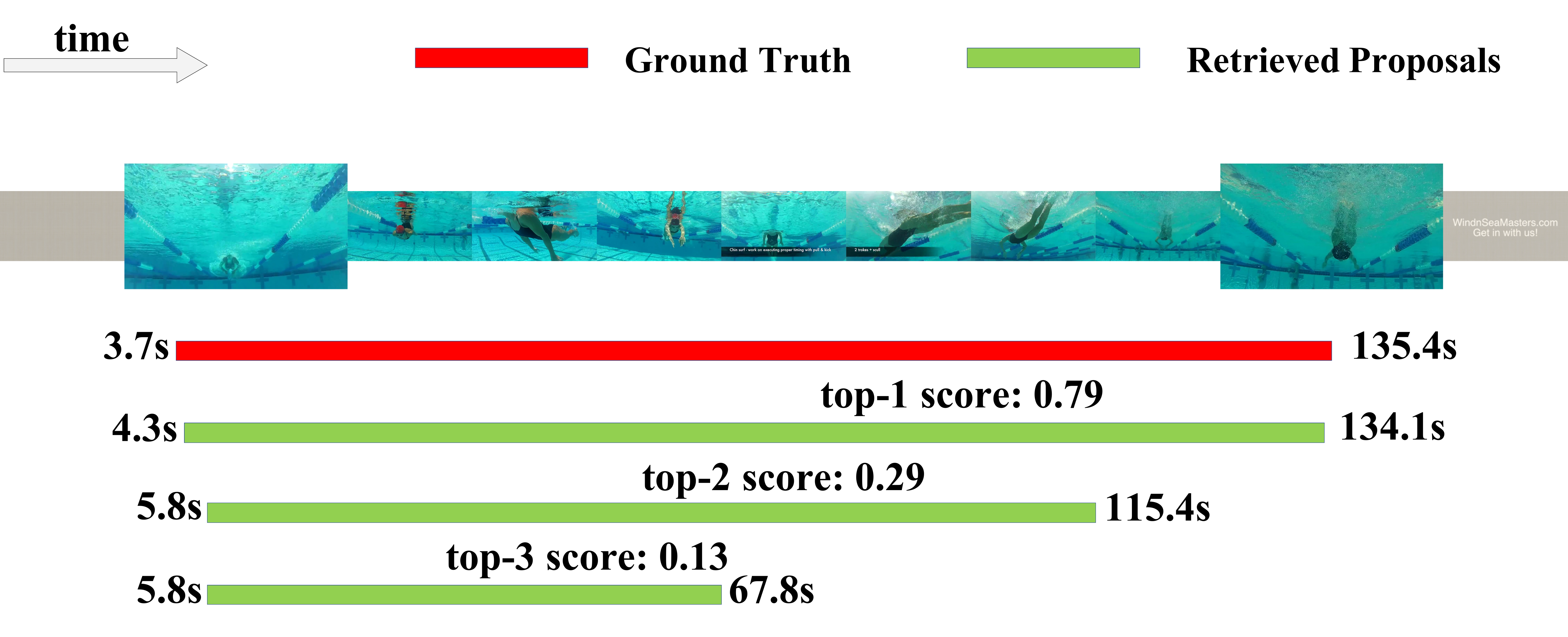}
			\label{fig:localization_b}
		}
		\quad
		\subfloat[]{
			\includegraphics[width=1 \columnwidth]{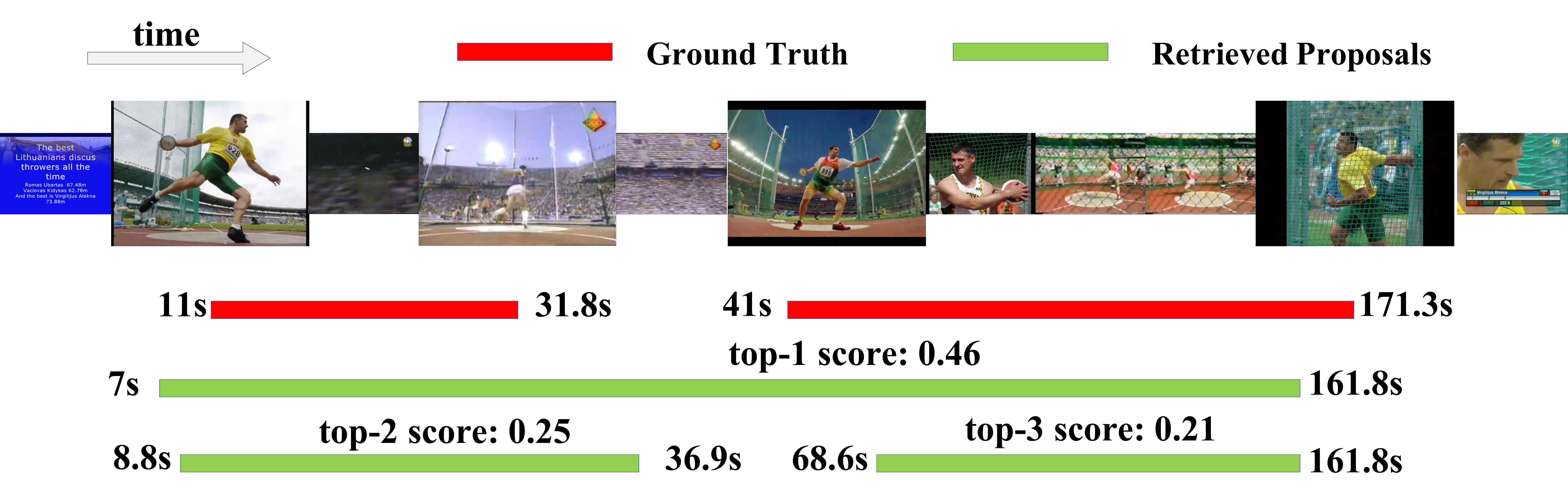}
			\label{fig:localization_c}
		}
		\hspace{2.2mm}
		\subfloat[]{
			\includegraphics[width=1 \columnwidth]{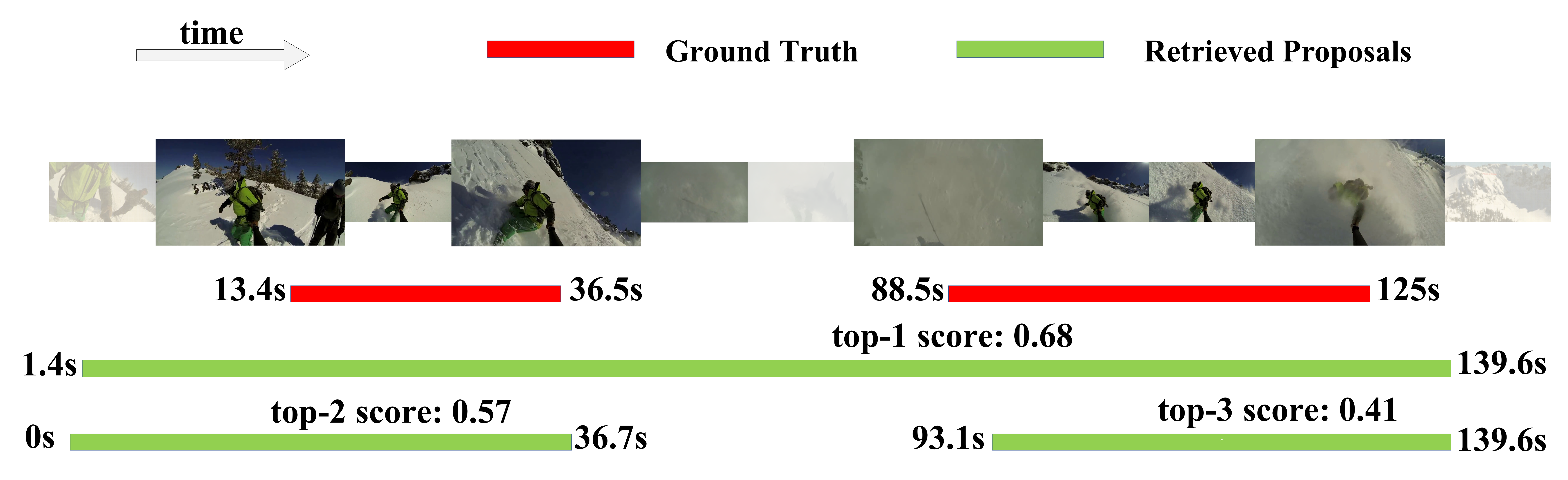}
			\label{fig:localization_d}
		}
		\caption{\textbf{Downstream action localization results}. We show our top-3 proposals and most of them align closely with the ground truth action instance. }
		\label{fig:localization_supp}
	\end{figure*}

	In Fig.~\ref{fig:localization_supp}, we show our top-3 action proposals and most of them align closely with the ground truth action instance. For both (c) and (d) in Fig.~\ref{fig:localization_supp}, we have proposals that can cover two instances in one shot, as well as individual proposal to align with each instance.

	\subsection{Feature visualization}
\begin{figure*}[t]
		\centering
		\subfloat[]{
			\includegraphics[width=1  \columnwidth]{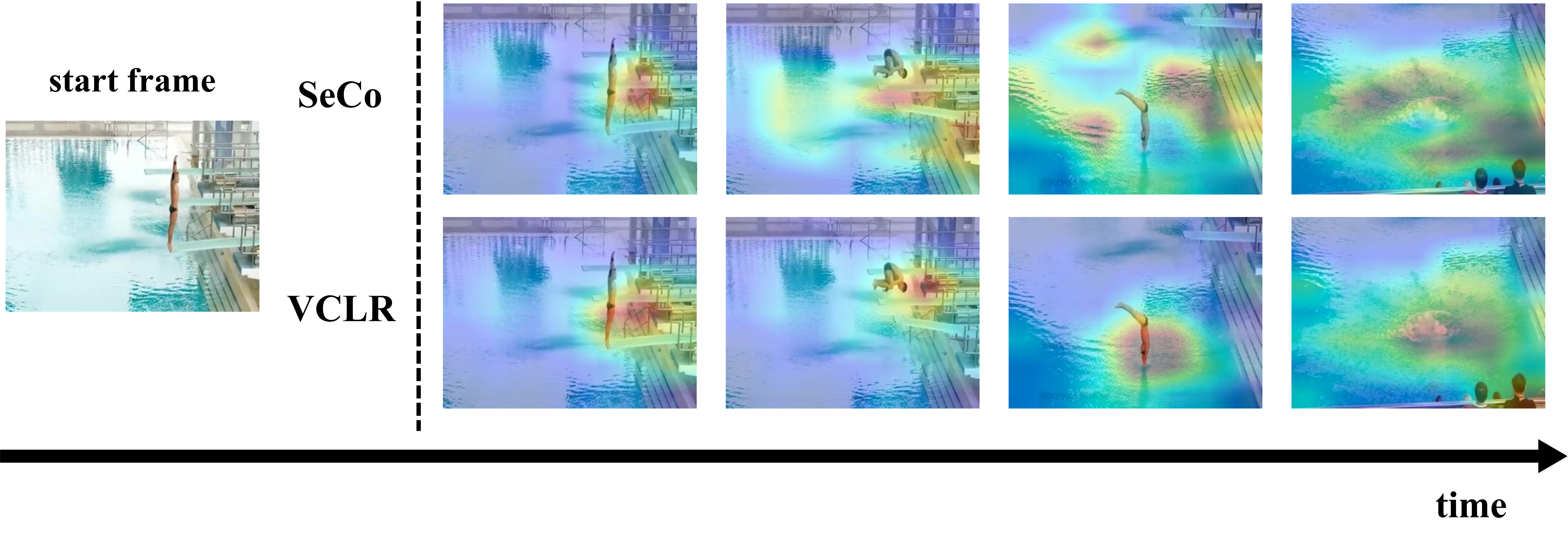}
			\label{fig:feat_k400}
		}
		\hspace{2.2mm}
		\subfloat[]{
			\includegraphics[width=1  \columnwidth]{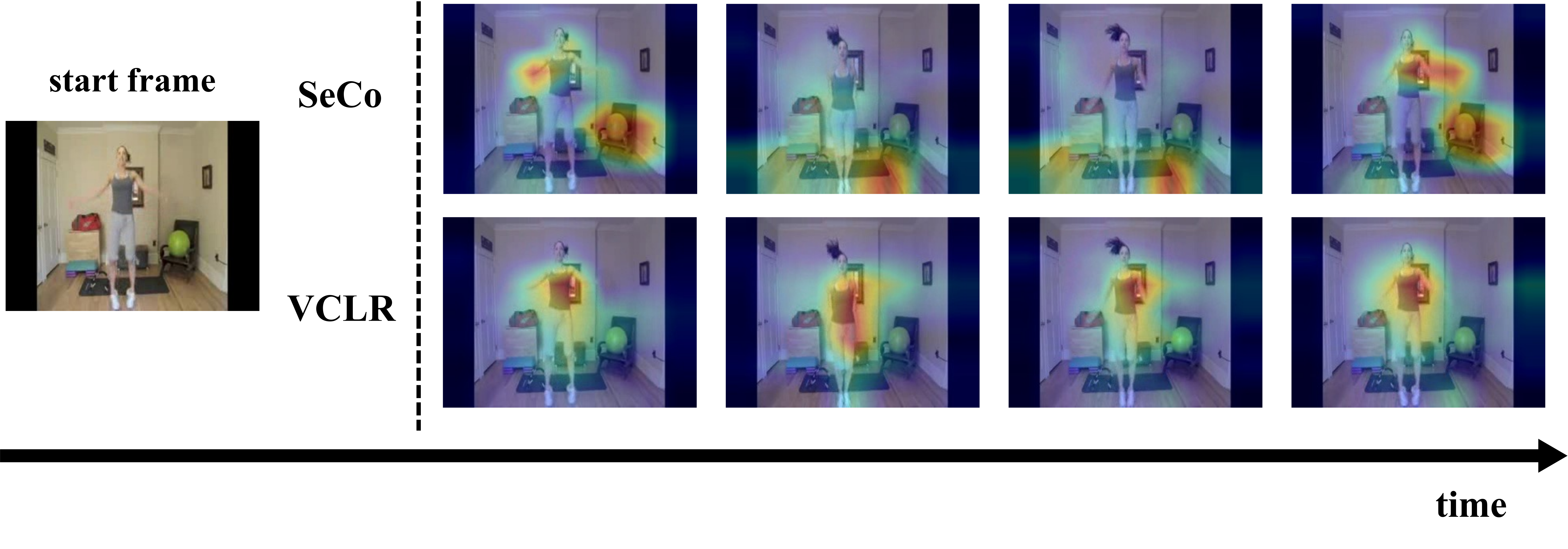}
			\label{fig:feat_ucf}
		}
		\quad
		\subfloat[]{
			\includegraphics[width=1 \columnwidth]{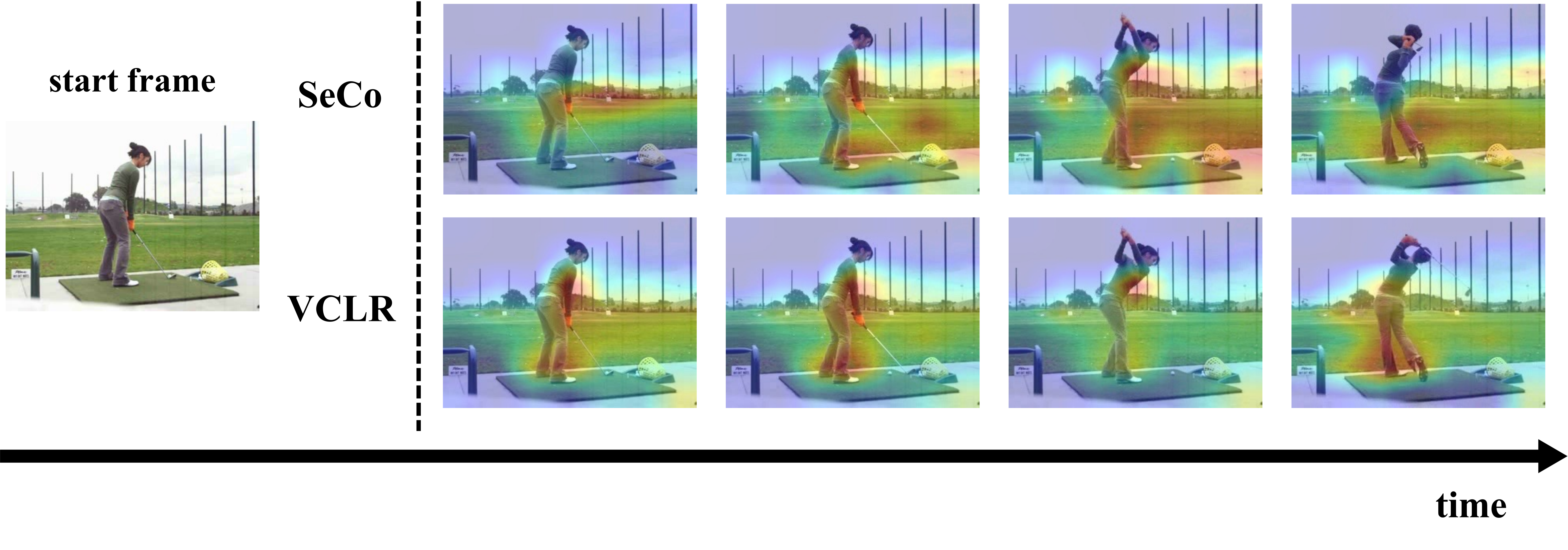}
			\label{fig:feat_hmdb}
		}
		\hspace{2.2mm}
		\subfloat[]{
			\includegraphics[width=1 \columnwidth]{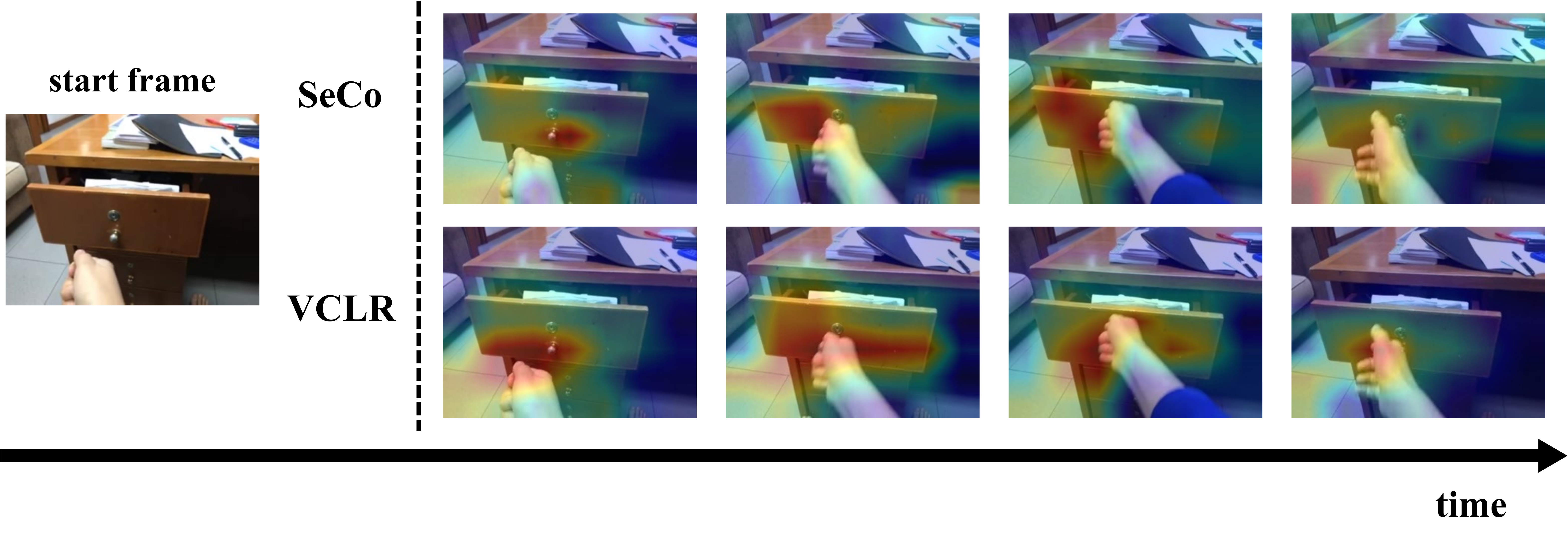}
			\label{fig:feat_sthv2}
		}
		\hspace{2.2mm}
		\subfloat[]{
			\includegraphics[width=1 \columnwidth]{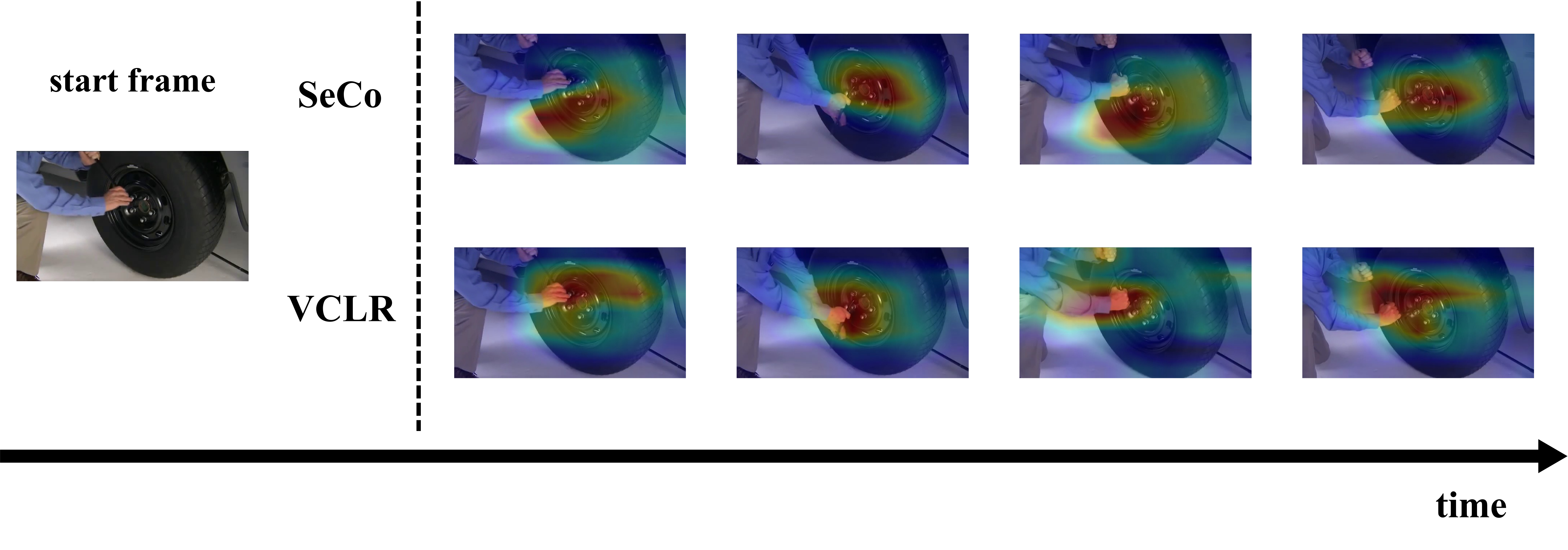}
			\label{fig:feat_anet}
		}
		\hspace{2.2mm}
		\subfloat[]{
			\includegraphics[width=1 \columnwidth]{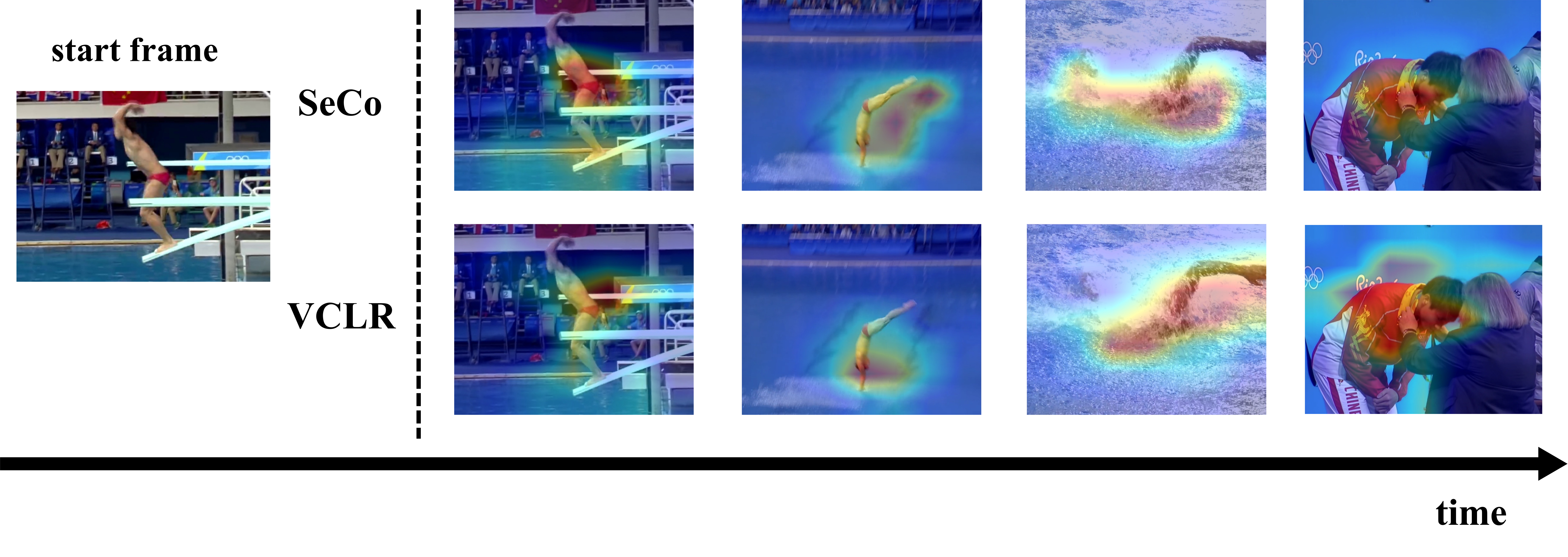}
			\label{fig:feat_unseen}
		}
		\caption{\textbf{More feature visualizations based on Grad-CAM}. (a) to (f) are examples from Kinetics400, UCF101, HMDB51, SthSthV2, ActivityNet, and unseen video respectively.}
		\label{fig:feat_supp}
	\end{figure*}

	We provide more feature visualizations in Fig.~\ref{fig:feat_supp}. We implement the Grad-Cam~\cite{selvaraju_iccv2017_gradcam} algorithm to generate attention maps with the output from conv5 layer of ResNet50. We select one example randomly from each dataset (Kinetics400, UCF101, HMDB51, SthSthV2 and ActivityNet) and also an unseen video. We compare our attention maps with SeCo~\cite{yao_aaai2021_seco} and see that VCLR is able to focus on the moving
	region of interest, while SeCo focuses more on the background.
	
\end{document}